\pdfoutput=1

\documentclass[11pt]{article}

\usepackage[final]{acl}

\usepackage{times}
\usepackage{latexsym}

\usepackage{lipsum}
\usepackage{booktabs}
\usepackage{amssymb}
\usepackage{amsmath}
\usepackage{enumitem}
\usepackage{pifont}
\usepackage{tcolorbox}
\usepackage{graphicx}
\usepackage{scalerel}
\usepackage{array}
\usepackage[normalem]{ulem}
\newcommand{\xmark}{\ding{55}}
\newcommand{\cmark}{\ding{51}}

\usepackage[frozencache,cachedir=.]{minted}

\usepackage{minted}
\setminted{
frame=lines,
framesep=2mm,
baselinestretch=1.2,
fontsize=\footnotesize
}

\newcommand{\checkmrk}{\scalerel*{\includegraphics{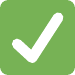}}{\textrm{\textbigcircle}}}
\newcommand{\cross}{\scalerel*{\includegraphics{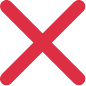}}{\textrm{\textbigcircle}}}

\definecolor{lightred}{HTML}{CC8685}
\definecolor{darkred}{HTML}{FF0000}

\usepackage[T1]{fontenc}

\usepackage[utf8]{inputenc}

\usepackage{microtype}

\usepackage{inconsolata}

\title{RAID: A Shared Benchmark for Robust Evaluation \\ of Machine-Generated Text Detectors}

\author{Liam Dugan$^1$, \hspace{0.25cm} Alyssa Hwang$^1$, \hspace{0.25cm} Filip Trhlik$^2$, \hspace{0.25cm} Josh Magnus Ludan$^1$\\\textbf{Andrew Zhu$^1$, \hspace{0.25cm} Hainiu Xu$^3$, \hspace{0.25cm} Daphne Ippolito$^4$, \hspace{0.25cm} Chris Callison-Burch$^1$} \\
University of Pennsylvania$^1$ \hspace{0.25cm} University College London$^2$\\ King's College London$^3$ \hspace{0.25cm} Carnegie Mellon University$^4$\\\ 
{\tt \normalsize \{ldugan, ahwang16, jludan, andrz, ccb\}@seas.upenn.edu}\\ \tt \normalsize hainiu.xu@kcl.ac.uk, filip.trhlik.21@ucl.ac.uk, daphnei@cmu.edu}

\begin{document}
\maketitle
\begin{abstract}
Many commercial and open-source models claim to detect machine-generated text with extremely high accuracy (99\% or more). However, very few of these detectors are evaluated on shared benchmark datasets and even when they are, the datasets used for evaluation are insufficiently challenging—lacking variations in sampling strategy, adversarial attacks, and open-source generative models. In this work we present RAID: the largest and most challenging benchmark dataset for machine-generated text detection. RAID includes over 6 million generations spanning 11 models, 8 domains, 11 adversarial attacks and 4 decoding strategies. Using RAID, we evaluate the out-of-domain and adversarial robustness of 8 open- and 4 closed-source detectors and find that current detectors are easily fooled by adversarial attacks, variations in sampling strategies, repetition penalties, and unseen generative models. We release our data\footnote{\url{https://github.com/liamdugan/raid}} along with a leaderboard\footnote{\url{https://raid-bench.xyz/leaderboard}} to encourage future research.
\end{abstract}

\section{Introduction}
Large Language Models (LLMs) have been able to fool humans into thinking their outputs are human-written for roughly four years \cite{dugan-etal-2020-roft, clark-etal-2021-thats}. In that short span of time we have seen LLM-generated text be used for targeted phishing attacks \cite{baki-etal-2017-scaling,hazell2023large}, mass spam and harassment \cite{weiss2019}, disinformation campaigns \cite{sharevski2023talking,spitale-etal-2023-disinformation}, and spurious scientific publication \cite{lund2023chatgpt}. In order to document and eventually mitigate such harms, we must develop robust automatic detectors of machine-generated text. 

Many exciting and inventive methods have been proposed in recent years for detecting generated text \cite{crothers-etal-2023-comprehensive}. However, when evaluating these methods, authors typically generate their own evaluation datasets and fail to test their models on shared resources---making it difficult to verify claims of accuracy and robustness. This has led to an erosion of trust in the efficacy of automatic detection methods and a generally fatalistic sentiment towards detection among researchers and practitioners \cite{sadasivan2023aigenerated}.

\begin{figure}
    \centering
    \includegraphics[width=\columnwidth]{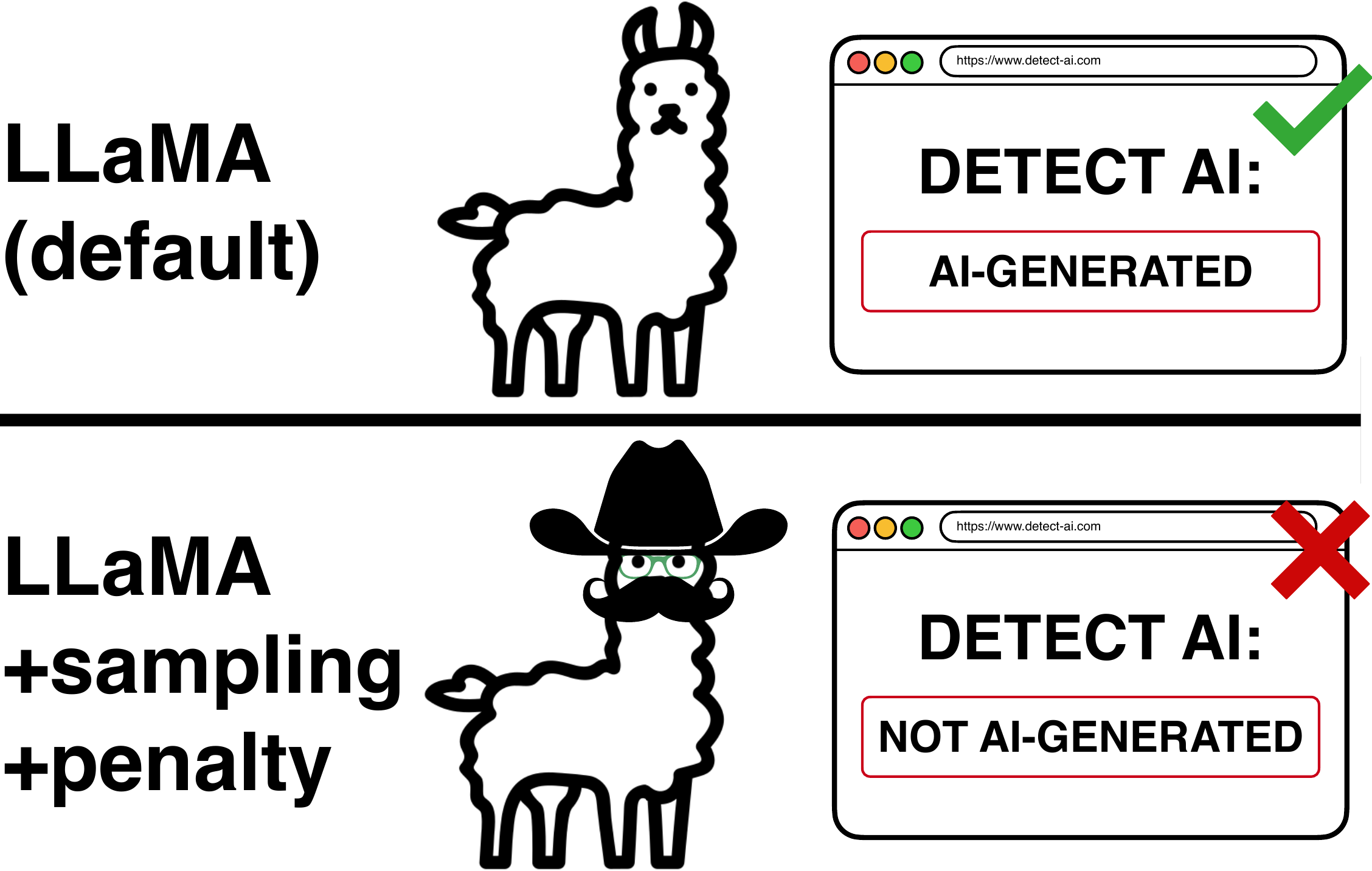}
    \caption{Detectors for machine-generated text are often highly performant on default model settings but fail to detect more unusual settings such as using random sampling with a repetition penalty.}
    \label{fig:overview}
\end{figure}

\begin{table*}[t]
    \small
    \centering
    \begin{tabular}{l|c|c|c|c|c|c}
    \toprule
    &&\textbf{Domain}&\textbf{Model}&\textbf{Sampling}&\textbf{Multilingual}&\textbf{Adversarial}\\
    \textbf{Name}&\textbf{Size}&\textbf{coverage?}&\textbf{coverage?}&\textbf{coverage?}&\textbf{coverage?}&\textbf{coverage?}\\
    \midrule
    
    TuringBench \cite{uchendu2021turingbench}&200k&\cross&\checkmrk&\cross&\cross&\cross\\ 
    RuATD \cite{Shamardina_2022}&215k&\checkmrk&\checkmrk&\cross&\cross&\cross\\
    HC3 \cite{guo2023close}&26.9k&\checkmrk&\cross&\cross&\checkmrk&\cross\\ 
    MGTBench \cite{he2023mgtbench}&2817&\checkmrk&\checkmrk&\cross&\cross&\checkmrk\\ 
    MULTITuDE \cite{macko-etal-2023-multitude}&74.1k&\cross&\checkmrk&\cross&\checkmrk&\cross\\ 
    AuText2023 \cite{sarvazyan2023overview}&160k&\checkmrk&\cross&\cross&\checkmrk&\cross\\ 
    M4 \cite{wang2023m4}&122k&\checkmrk&\checkmrk&\cross&\checkmrk&\cross\\ 
    CCD \cite{wang2023evaluating}&467k&\cross&\cross&\cross&\checkmrk&\checkmrk\\ 
    IMDGSP \cite{mosca-etal-2023-distinguishing}&29k&\cross&\checkmrk&\cross&\cross&\cross\\ 
    HC-Var \cite{xu2023generalization}&145k&\checkmrk&\cross&\cross&\cross&\cross\\ 
    HC3 Plus \cite{su2024hc3}&210k&\checkmrk&\cross&\cross&\checkmrk&\cross\\ 
    MAGE \cite{li2024mage}&447k&\checkmrk&\checkmrk&\cross&\cross&\cross\\ 
    \midrule
    \textbf{RAID (Ours)}&6.2M&\checkmrk&\checkmrk&\checkmrk&\cross&\checkmrk\\
    \bottomrule
    \end{tabular}
    \caption{A comparison of the publicly available sources of generated text. Our provided dataset is the only one that contains a diverse selection of domains, sampling strategies, and adversarial attacks across recent generative models.}
    \label{tab:all-datasets}
\end{table*}

To combat this trend, in this work, we introduce the Robust AI Detection (RAID) benchmark. RAID is the largest and most challenging benchmark of generated text ever released, consisting of 6M+ generations spanning 11 generators, 8 domains, 11 adversarial attacks, and 4 decoding strategies. Using RAID, we benchmark 12 detectors (8 open- and 4 closed-source). We find that detectors have difficulty generalizing to unseen models and domains and that simple changes such as changing the sampling strategy, adding a repetition penalty, and adversarially modifying text lead to marked decreases in performance.

\section{Related Work}
In Table \ref{tab:all-datasets} we show a comparison between RAID and other publicly available sources of generated text. Among these, the most similar work to ours is \citet{li2024mage}, who create a dataset of 447k generations from 7 language model families across 10 domains to study detector robustness. Other resources typically focus on particular sub-areas such as multilingual text \cite{macko-etal-2023-multitude,wang2023m4}, code \cite{wang2023evaluating}, question-answering \cite{guo2023close,xu2023generalization,su2024hc3}, and scientific papers \cite{mosca-etal-2023-distinguishing}. Additionally, shared tasks such as AuTextTification \cite{sarvazyan2023overview} and RuATD \cite{Shamardina_2022} have provided datasets and encouraged centralized evaluation and competition. 
While many shared resources do well at covering multiple generative models and domains, few include adversarial attacks and none include variation in decoding strategy---frequently even failing to list the strategy used. These datasets are insufficiently challenging and promote the inflated reports of detector accuracy. 

Another way to evaluate robustness is through a small-scale comparative study. In these studies, one aspect of the generated text is varied and detector accuracy is compared across the variations. Such studies have shown that detectors lack robustness to unseen generative models \cite{stiff-and-johansson-2022,xiaopu2023zeroshot,chakraborty-etal-2023-counter}, domains \cite{pagnoni-etal-2022-threat,pu-etal-2023-deepfake,rodriguez-etal-2022-cross}, decoding strategies \cite{ippolito-etal-2020-automatic,solaiman-etal-2019}, prompts \cite{koike2023prompt,kumarage2023reliable,lu2023large}, repetition penalties \cite{fishchuk2023efficient}, and human edits \cite{gao2024llmasacoauthor}. 

Similar work specializing in adversarial robustness has shown that detectors are vulnerable to homoglyph attacks \cite{gagiano-etal-2021-robustness, wolff2020attacking, macko2024authorship}, whitespace insertion \cite{cai2023evade}, sentiment and factual alterations \cite{bhat-parthasarathy-2020-effectively}, paraphrase attacks \cite{krishna2023paraphrasing,sadasivan2023aigenerated}, and synonym replacement \cite{kulkarni2023exploring,pu-etal-2023-deepfake}. Our work builds on this foundation and synthesizes many elements of the robustness literature into one singular systematic benchmark study. 

\begin{figure*}
    \centering
    \includegraphics[width=\textwidth]{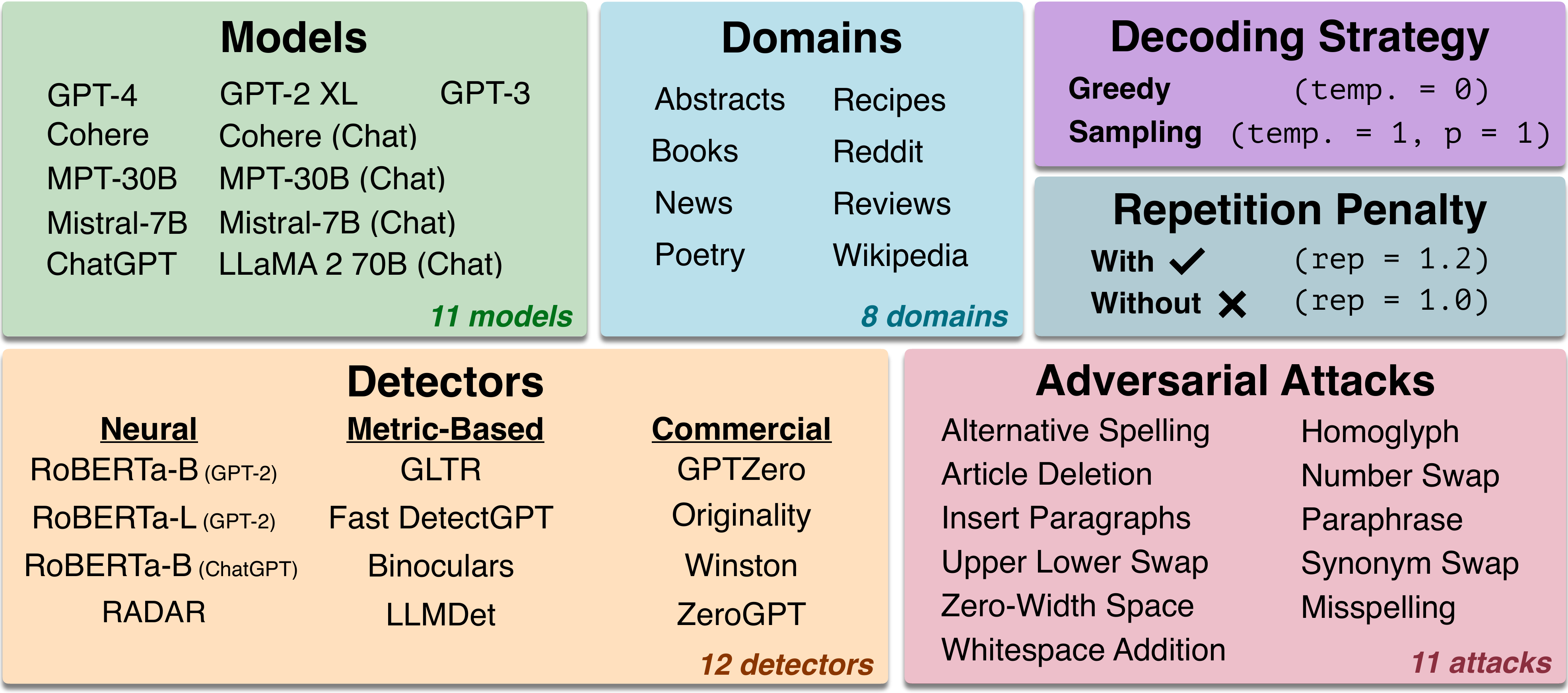}
    \caption{An overview of the structure of the RAID dataset. We generate 2,000 continuations for every combination of domain, model, decoding, penalty, and adversarial attack. This results in roughly 6.2 million generations for testing. We then evaluate each detector on all pieces of generated text in the dataset.}
    \label{fig:raid-figure}
\end{figure*}

\section{Dataset Creation}
\subsection{Overview}
In Figure \ref{fig:raid-figure}, we illustrate the components of the RAID dataset. To create RAID, we first sample roughly 2,000 documents of human-written text from each of our 8 target domains (\S\ref{sec:domains}). For each document, we create a corresponding generation prompt using a template such as ``Write a recipe for \{title\}'' (\S\ref{sec:prompts}). We then generate one output for each of our 11 models (\S\ref{sec:models}), 4 decoding strategies (\S\ref{sec:decoding-strategies}), and 11 adversarial attacks (\S\ref{sec:adversarial-attacks}). The RAID dataset consists of over 6M generations, the largest dataset of generated text to date.

\subsection{Domains}
\label{sec:domains}
Since different domains have been shown to induce LLMs to make diverse errors \citep{dugan-etal-2023-roft}, we prioritized domains that were both at high risk for abuse and were diverse and challenging. Our sources require factual knowledge (News, Wikipedia), generalization and reasoning (Abstracts, Recipes), creative and conversational skills (Reddit, Poetry), and knowledge of specific media (Books, Reviews). To avoid contamination, most of our human-written documents are taken from publicly available pre-2022 datasets (see Appendix \ref{app:domains}).

\subsection{Prompts}
\label{sec:prompts}
We prompt our generators in a zero-shot fashion using ``Chat'' templates for models fine-tuned on dialogue and ``Non-Chat'' templates for continuation models. Each prompt is nearly the same, with the exception of a ``\{title\}'' field that is dynamically replaced with the title of the corresponding human-written text (see Table \ref{tab:prompt-templates}). Unlike previous work \cite{verma2023ghostbuster, xu2023generalization}, we intentionally avoid biasing the language model towards a particular length or generation style to better match our expectations of real-world scenarios. We engineered our prompts over multiple rounds to minimize degenerate repetition, unhelpful generation, and meta-commentary across all models (see Appendix \ref{app:prompts}).

\subsection{Models}
\label{sec:models}
We carefully chose a set of models that were maximally distinct from each other, offering us the widest range and variability of generated text. We focused on varying model sizes, open/closed source, and chat/completion style. Following work by \citet{sarvazyan-etal-2023-supervised}, we select largest model from each model family and exclude third-party fine-tuned variants of base models in favor of the base models themselves. In total, we used four GPT models (GPT2, GPT3, GPT4, ChatGPT), three open-source models and their chat variants (Mistral 7B, MPT 30B, LLaMA 2 70B), and the Cohere \textit{command} and \textit{chat} model (see Appendix \ref{app:models}). 

\begin{table}[t]
\centering 
\small
\begin{tabular}{l|p{0.7\linewidth}}
\toprule
\textbf{Chat}&Write the abstract for the academic paper titled "\{title\}".\\
\midrule
\textbf{Non-Chat}&The following is the full text of the abstract for a research paper titled "\{title\}" from arxiv.org:\\
\bottomrule
\end{tabular}
\caption{The ``Chat'' and ``Non-Chat'' templates used for generation in the Abstracts domain. The ``\{title\}'' field is dynamically filled in with the title of the human-written document at generation time.}
\label{tab:prompt-templates}
\end{table}

\subsection{Decoding Strategies}
\label{sec:decoding-strategies}
The decoding strategy determines how tokens are selected from the language model's probability distribution. Previous work has shown that greedy decoding (i.e. selecting the most likely token at each time step) reduces the diversity of text and makes it easier to detect while sampling directly from the language model output distribution shows the opposite effect \citep{ippolito-etal-2020-automatic}. Based on these findings, we generate two outputs per prompt, one with greedy decoding and the other with fully random sampling.

We also generate two additional outputs with \citet{keskar2019ctrl}'s repetition penalty when available. This penalty works by down-weighting the probability of tokens that have previously appeared in the context window by some multiplicative factor $\theta$, resulting in less repetitive output. We are the first to evaluate this penalty for detection at a large scale and show that it significantly reduces the detectability of outputs. Following \citet{keskar2019ctrl}, we use $\theta=1.2$ for our experiments.\footnote{Only open-source models provide access to a repetition penalty. OpenAI and Cohere instead offer slightly different ``frequency'' and ``presence'' penalties (see Appendix \ref{app:repetition_penalty}). Thus we vary repetition penalty only for open-source models.}

\subsection{Adversarial Attacks}
\label{sec:adversarial-attacks}
When selecting adversarial attacks, we assume that our adversary has exactly one query and no knowledge of the detector. Thus, we include the following 11 black-box, query-free attacks as opposed to gradient-based methods:
\begin{enumerate}[noitemsep]
    \item \textbf{Alternative Spelling}: Use British spelling
    \item \textbf{Article Deletion}: Delete (`the', `a', `an')
    \item \textbf{Add Paragraph}: Put \texttt{\textbackslash n\textbackslash n} between sentences
    \item \textbf{Upper-Lower}: Swap the case of words
    \item \textbf{Zero-Width Space}: Insert the zero-width space \texttt{U+200B} every other character
    \item \textbf{Whitespace}: Add spaces between characters
    \item \textbf{Homoglyph}: Swap characters for alternatives that look similar, e.g. e $\rightarrow$ \texttt{e} (\texttt{U+0435})
    \item \textbf{Number}: Randomly shuffle digits of numbers
    \item \textbf{Misspelling}: Insert common misspellings 
    \item \textbf{Paraphrase}: Paraphrase with the fine-tuned T5-11B model from \citet{krishna2023paraphrasing}
    \item \textbf{Synonym}: Swap tokens with highly similar BERT \cite{devlin-etal-2019-bert} candidate tokens   
\end{enumerate}

Following recommendations from \citet{dyrmishi-etal-2023-humans}, we manually reviewed our data to ensure that adversarial attacks were inconspicuous. Thus, for each attack only a small percentage of the total available mutations were applied. For details on mutation percentages for each attack as well as other implementation details, see Appendix \ref{app:adversarial}.

\subsection{Post-Processing}
After all generations were completed, we removed prompts and left only the generated output. We then filtered out failed generations and balanced the dataset such that each human-written document has exactly one corresponding generation per model, decoding strategy, and adversarial attack.

\begin{figure}
    \centering
    \includegraphics[width=\columnwidth]{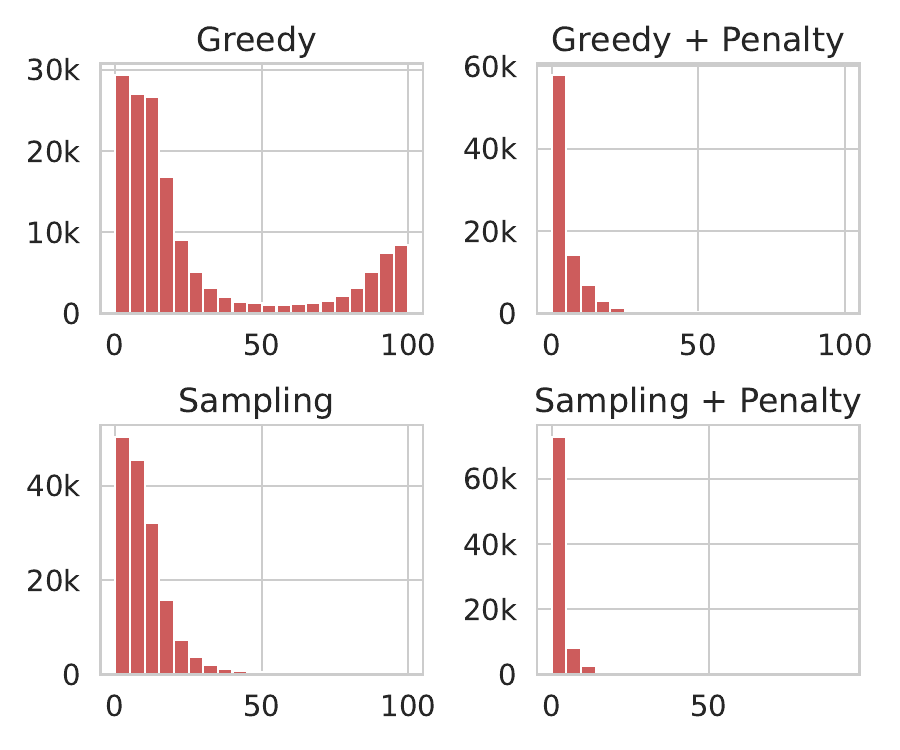}
    \caption{Histogram of examples (y-axis) grouped by their repetitiveness measured via SelfBLEU score (x-axis). We see that both random sampling and repetition penalty greatly reduce repetitiveness for all models.}
    \label{fig:selfbleu}
\end{figure}

\begin{table}[t]
    \small
    \centering
    \begin{tabular}{l|c|c|c|c|c}
    \toprule
    &\textbf{Num.}&&\textbf{Self-}&\textbf{PPL}&\textbf{PPL}\\
    \textbf{Model}&\textbf{Gens}&\textbf{Toks}&\textbf{BLEU}&\textbf{-L7B}&\textbf{-G2X}\\
    \midrule
    \textbf{Human}&14971&378.5&7.64&9.09&21.2\\
    \midrule
    \textbf{GPT 2}&59884&384.7&23.9&8.33&8.10\\
    \textbf{GPT 3}&29942&185.6&13.6&3.90&8.12\\
    \textbf{ChatGPT}&29942&329.4&10.3&3.39&9.31\\
    \textbf{GPT 4}&29942&350.8&9.42&5.01&13.4\\
    \textbf{Cohere}&29942&301.9&11.0&5.67&23.7\\
    \textit{(+ Chat)}&29942&239.0&11.0&4.93&11.6\\
    \textbf{Mistral}&59884&370.2&19.1&7.74&17.9\\
    \textit{(+ Chat)}&59884&287.7&9.16&4.31&10.3\\
    \textbf{MPT}&59884&379.2&22.1&14.0&66.9\\
    \textit{(+ Chat)}&59884&219.2&5.39&7.06&56.3\\
    \textbf{LLaMA}&59884&404.4&10.6&3.33&9.76\\
    \midrule
    \textbf{Total}&509k&323.4&13.7&6.61&23.8\\
    \bottomrule
    \end{tabular}
    \caption{Statistics for the generations in the base dataset without adversarial attacks. \textbf{PPL-L7B} refers to mean perplexity according to LLaMA 7B and \textbf{PPL-G2X} refers to mean perplexity according to GPT 2 XL.}
    \label{tab:dataset-statistics}
\end{table}

\section{Dataset}
\subsection{Statistics}
The non-adversarial portion of the RAID dataset consists of 509,014 generations and 14,971 human-written documents for a total of 6,287,820 texts when including adversarial attacks. On average, models are more repetitive than humans as measured by Self-BLEU \cite{zhu2018texygen} and typically generate shorter passages (see Table \ref{tab:dataset-statistics}). The mean perplexity is lower for models than humans, according to LLaMA 7B and GPT 2 XL.%

\subsection{Release Structure}
\label{sec:release_structure}
To accompany the RAID dataset we also release an official public leaderboard\footnote{\url{https://raid-bench.xyz/leaderboard}} which will host the results from our analysis alongside other detector results submitted by public contributors. The leaderboard is split up into two sections---one for those who self-report having trained on the RAID dataset and one for those who do not. This is important to ensure that a clear distinction is made between detectors that are generalizing to out-of-domain data and those that are not. 

To ensure fair competition, 10\% of the RAID dataset is released without labels for use as the official hidden test set. We provide scripts to easily run detectors on this test set and calculate accuracy with respect to the hidden labels. Users can then submit their outputs to the leaderboard via a pull request (see Appendix \ref{app:leaderboard}). We hope that this infrastructure encourages more comparison and shared evaluation of detectors.

\subsection{RAID-extra}
In addition to the over 6M+ generations in the core RAID train and test sets, we also release ``RAID-extra'', an additional dataset consisting of 2.3M generations from three extra domains not included in the main benchmark: Python Code, Czech News, and German News. In Appendix \ref{app:multilingual} we report evaluation results from our 12 detectors on RAID-extra and show that metric-based detectors perform surprisingly well on these types of unusual domains. 

RAID-extra is the largest and most challenging dataset of generated code and multilingual text ever released. We hope it will be of value to the academic community.

\section{Detectors}
\label{sec:detectors}
\subsection{Detector Selection} 
We evaluate detectors from three categories: neural, metric-based, and commercial. Neural detectors typically involve fine-tuning a pre-trained language model such as RoBERTa \cite{liu2019roberta} while metric-based detectors typically compute some metric using the output probabilities of an existing generative model. 
In contrast, commercial detectors tend to provide some documentation of their performance but disallow direct access to the models. We tested the following:

\renewcommand{\theenumi}{\textbf{(\roman{enumi})}}
\begin{enumerate}[noitemsep]
    \item \textbf{Neural}: RoBERTa-Base (GPT2), RoBERTa-Large (GPT2), RoBERTa-Base (ChatGPT), RADAR
    \item \textbf{Metric-Based}: GLTR, Binoculars, Fast DetectGPT, LLMDet
    \item \textbf{Commercial}: GPTZero, Originality, Winston, ZeroGPT
\end{enumerate}

Unlike \citet{li2024mage}, we do not train our own neural models on our dataset because we wish to investigate the generalization ability of off-the-shelf detectors. For the metric-based detectors, we chose to use the default generative model in each repository to emulate the most realistic use-case.\footnote{See Appendix \ref{app:detectors} for more details on each detector tested}

\subsection{Detector Evaluation}
Detectors work by taking in a sequence of tokens and outputting a scalar score. In order to convert this score to a binary prediction, we must select a scalar threshold $\tau$ such that if the score $s \geq \tau$ the sequence is predicted to be machine-generated. 

In our work, we select a threshold for each model such that the resulting false positive rate of the detector is 5\%. In practical terms, accuracy at a fixed FPR of 5\% represents how well each detector identifies machine-generated text while only mis-classifying 5\% of human-written text. 
Our work is one of the first shared resources to fix and disclose FPR, following the rise of this evaluation paradigm in recent robustness research \citep{hans2024spotting, krishna2023paraphrasing, soto2024fewshot}.\footnote{See Appendix \ref{app:accuracy_vs_f1} for a discussion of why we chose this paradigm instead of the traditional precision/recall/F1 score}

\begin{table}[t]
    \small
    \centering
    \begin{tabular}{l|c|c|c|c}
    \toprule
    &$\tau$=0.25&$\tau$=0.5&$\tau$=0.75&$\tau$=0.95\\
    \midrule
    R-B GPT2&8.71\%&6.59\%&5.18\%&3.38\%\\
    R-L GPT2&6.14\%&2.91\%&1.46\%&0.25\%\\
    R-B CGPT&21.6\%&15.8\%&15.1\%&10.4\%\\
    RADAR&7.48\%&3.48\%&2.17\%&1.23\%\\
    \midrule
    GLTR&100\%&99.3\%&21.0\%&0.05\%\\
    F-DetectGPT&47.3\%&23.2\%&13.1\%&1.70\%\\
    LLMDet&97.9\%&96.0\%&92.0\%&75.3\%\\
    Binoculars&0.07\%&0.00\%&0.00\%&0.00\%\\
    \midrule
    GPTZero&0.03\%&0.00\%&0.00\%&0.00\%\\
    Originality&0.47\%&0.25\%&0.17\%&0.07\%\\
    Winston&0.75\%&0.55\%&0.38\%&0.21\%\\
    ZeroGPT&1.71\%&1.42\%&1.21\%&0.90\%\\
    \bottomrule
    \end{tabular}
    \caption{False Positive Rates for detectors on RAID at naive choices of threshold ($\tau$). We see that, for open-source detectors, thresholding naively results in unacceptably high false positive rates.}
    \label{tab:fpr_results}
\end{table}

\section{Findings}
\paragraph{Finding 1: Default False Positive Rates (FPRs) of open-source detectors are dangerously high}
When applying a detector to a piece of text, it is important to decide on a threshold ($\tau$) to use for binary classification. The principled way to determine this is to use a set of in-domain data to calibrate the threshold to a specific point along the receiver-operating curve. However, this is cumbersome and requires a set of readily accessible human-written data. Thus, in practice it is common to simply use some seemingly sensible default value (such as 0.5) for the threshold without further investigation.

In Table \ref{tab:fpr_results} we report the false positive rates of our detectors for various commonly chosen thresholds. We see that open-source detectors, especially metric-based detectors, exhibit dangerously high false positive rates when using these naive thresholds. On the contrary, closed-source detectors seem to be calibrated fairly well, with none having an FPR above 1.7\%. Given this result, we advise practitioners to take care not to use naive-yet-sensible values for their detectors and instead calibrate detectors on in-domain data before using them. 

Following this advice, for the remainder of the Findings section, we will be exclusively using thresholds that were calibrated to a FPR of 5\% on the human-written portion of the RAID dataset (see Appendix \ref{app:threshold_tuning}).

\begin{figure}
    \centering
    \includegraphics[width=\columnwidth]{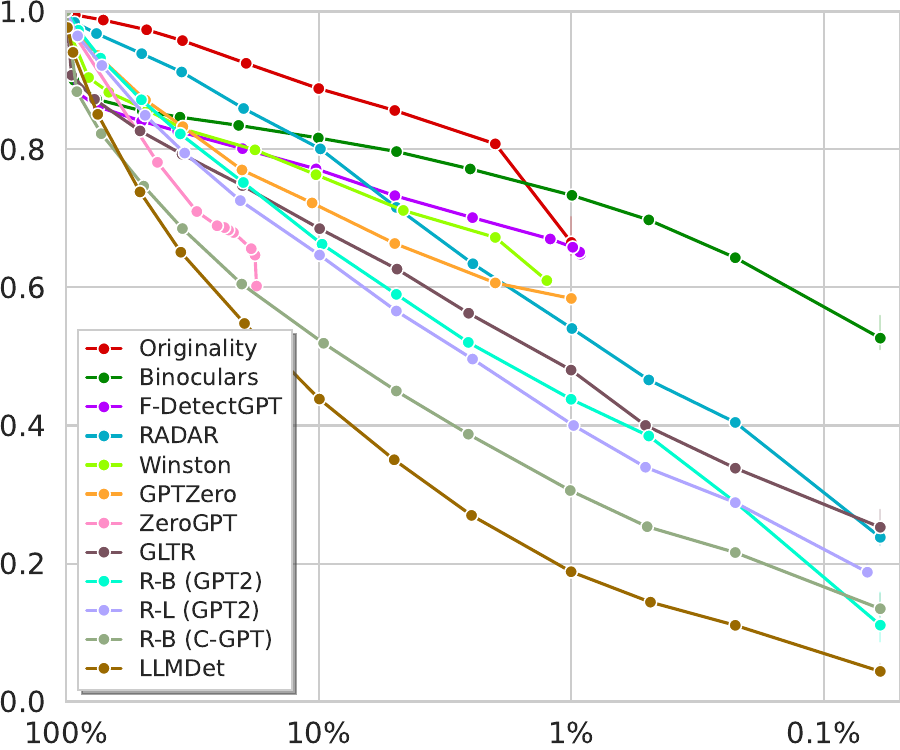}
    \caption{Detection accuracy (y-axis) vs. False Positive Rate (x-axis) for all detectors. We see that Binoculars works significantly better than other detectors at low FPR and that few detectors can operate at FPR<1\%.}
    \label{fig:fpr_lineplot}
\end{figure}

\begin{figure}[t]
    \centering
    \includegraphics[width=\columnwidth]{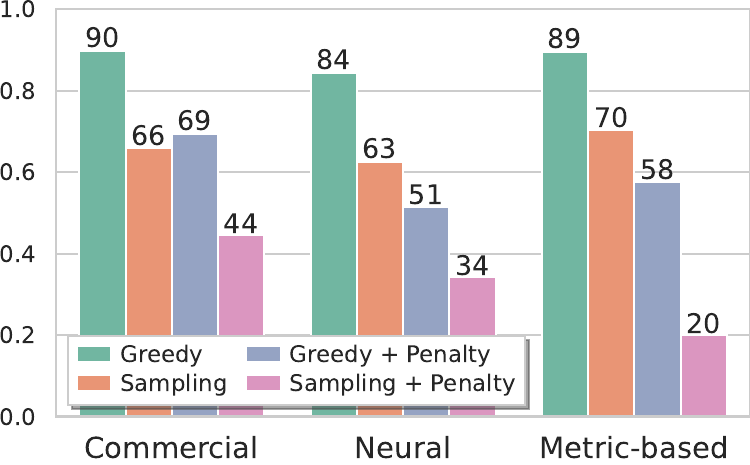}
    \caption{Accuracy at FPR=5\% (y-axis) vs. decoding strategy across our three detector classes on non-adversarial text. We see that repetition penalty greatly reduces accuracy in both greedy decoding and sampling.}
    \label{fig:decrep_barplot}
\end{figure}

\begin{figure*}[t]
\includegraphics[width=\textwidth]{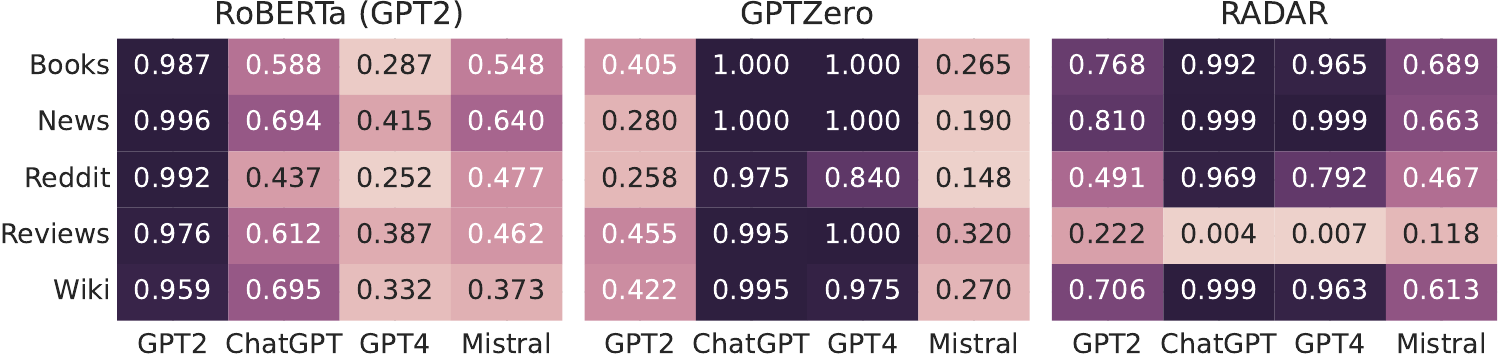}
    \caption{Heatmap measuring the accuracy of the RoBERTa-Large GPT2, GPTZero, and RADAR detectors across models and domains. We see a clear bias towards domains and models that the detectors have trained on.}
    \label{fig:heatmap}
\end{figure*}

\begin{table*}[t]
    \small
    \centering
    \begin{tabular}{l|c|c|c|c|c|c|c|c|c|c|c|c}
    \toprule 
    &\multicolumn{8}{c|}{\textbf{Open-Source}}&\multicolumn{4}{c}{\textbf{Closed-Source}}\\
    &\multicolumn{4}{c|}{\textbf{Chat Models}}&\multicolumn{4}{c|}{\textbf{Non-Chat Models}}&\multicolumn{2}{c|}{\textbf{Chat Models}}&\multicolumn{2}{c}{\textbf{Non-Chat Models}}\\
    &\multicolumn{4}{c|}{(llama-c, mistral-c, mpt-c)}&\multicolumn{4}{c|}{(mistral, mpt, gpt2)}&\multicolumn{2}{c|}{(c-gpt, gpt4, cohere)}&\multicolumn{2}{c}{(cohere, gpt3)}\\
    \midrule
    Dec. Strategy&\multicolumn{2}{c|}{greedy}&\multicolumn{2}{c|}{sampling}&\multicolumn{2}{c|}{greedy}&\multicolumn{2}{c|}{sampling}&greedy&sampling&greedy&sampling\\
    \midrule
    Rep. Penalty?&\xmark&\cmark&\xmark&\cmark&\xmark&\cmark&\xmark&\cmark&\xmark&\xmark&\xmark&\xmark\\
    \midrule
    R-B GPT2&84.1&52.3&77.9&26.2&98.6&44.1&60.5&35.4&70.9&41.7&65.1&52.5\\
    R-L GPT2&79.7&41.1&71.4&19.5&98.5&43.0&67.2&53.4&61.4&34.7&61.1&48.6\\
    R-B CGPT&80.2&63.3&75.0&39.3&53.3&26.4&14.9&1.7&59.1&38.1&46.5&39.0\\
    RADAR&88.8&77.4&85.6&66.4&91.8&63.8&48.3&31.8&81.6&75.3&72.2&67.7\\
    \midrule
    GLTR&89.8&67.5&83.9&38.3&99.6&56.9&44.5&0.5&80.7&54.3&75.6&63.7\\
    F-DetectGPT&98.6&74.5&96.2&40.5&97.8&56.1&79.7&0.6&96.0&74.1&93.8&86.3\\
    LLMDet&55.5&30.2&47.5&16.5&74.8&27.0&38.4&3.7&35.8&18.5&40.0&32.9\\
    Binoculars&\textbf{99.9}&86.6&\textbf{99.7}&60.6&\textbf{99.9}&62.3&72.4&0.6&\textbf{99.2}&92.1&\textbf{99.0}&\textbf{95.0}\\
    \midrule
    GPTZero&98.8&\textbf{93.7}&98.4&\textbf{82.5}&74.7&34.6&9.4&4.8&92.3&88.5&60.6&53.4\\
    Originality&98.6&86.3&97.7&72.5&\textbf{99.9}&\textbf{64.1}&\textbf{89.0}&\textbf{51.2}&96.8&89.0&91.7&85.4\\
    Winston&97.2&90.1&96.6&78.3&68.2&49.0&29.5&11.3&96.1&\textbf{93.7}&73.2&68.1\\
    ZeroGPT(*)&95.4&80.7&90.5&54.9&85.1&57.2&16.0&0.3&92.1&65.8&83.4&72.7\\
    \bottomrule
    \end{tabular}
    \caption{Accuracy Score at FPR=5\% for all detectors across model groups and sampling strategies. Asterisks (*) indicate that the detector was unable to achieve the target FPR. We see that random sampling with a repetition penalty consistently makes output generations very difficult to detect, especially for open-source non-chat models.}
    \label{tab:decoding_full_results}
\end{table*}

\paragraph{Finding 2: Detector accuracy varies substantially depending on target False Positive Rate}
In Figure \ref{fig:fpr_lineplot} we report the results of an experiment where we varied the classification threshold and plotted detector accuracy vs. false positive rate. We found that our detectors were capable of achieving the high accuracies cited in many viral reports, but only at similarly high FPR. 
Some detectors failed to achieve the lowest FPR we tested, plateauing at 16.9\% (ZeroGPT), 0.88\% (FastDetectGPT), and 0.62\% (Originality). Most detectors dropped steeply as FPR decreased from 100\%, but Binoculars \citep{hans2024spotting} was particularly strong at low FPR. Since detector accuracy varies so much with FPR, explicitly calibrating and reporting FPR is crucial for comparable, informative, and reproducible detection studies.

\paragraph{Finding 3: Repetition penalty drastically hurts accuracy for all detectors}
As shown in Figure \ref{fig:decrep_barplot}, we observe a consistent pattern across detectors, detector categories, generators, and domains: adding a repetition penalty decreases accuracy by up to 32 points regardless of decoding strategy. The decoding strategy matters as well. Detectors in each category, across domains and generators, perform substantially better on greedy decoding than random sampling, even when taking repetition penalty into account.

This pattern is especially concerning because past studies have largely overlooked variations in decoding strategy when evaluating detectors. Furthermore, none have reported results on repetition penalty before our study. Since sampling with repetition penalty results in text that often sounds more human-like \citep{keskar2019ctrl}, exposing these patterns is critical for reliable detection. 

There are many possible penalties and decoding strategies one could use when generating text such as contrastive decoding \citep{li-etal-2023-contrastive}, eta and epsilon decoding \citep{hewitt-etal-2022-truncation}, and typical sampling \citep{meister-etal-2023-locally}---all of which are likely to reduce detector accuracy. We highly encourage robustness studies and open evaluations to investigate how well detectors can generalize to alternative generation settings, especially if generative models are already trained on such outputs.

\begin{table*}[t]
    \small
    \centering
    \begin{tabular}{l|c|c|c|c|c|c|c}
    \toprule
    &\textbf{None}&\textbf{Paraphrase}&\textbf{Synonym}&\textbf{Misspelling}&\textbf{Homoglyph}&\textbf{Whitespace}&\textbf{Delete Articles}\\
    \midrule
    R-L GPT2&56.7&72.9 ({\color{blue}+16.2})&79.4 ({\color{blue}+22.7})&39.5 ({\color{darkred}-17.2})&21.3 ({\color{darkred}-35.4})&40.1 ({\color{darkred}-16.6})&33.2 ({\color{darkred}-23.5})\\
    RADAR&70.9&67.3 ({\color{lightred}-3.6})&67.5 ({\color{lightred}-3.4})&69.5 ({\color{lightred}-1.4})&59.3 ({\color{darkred}-11.6})&66.1 ({\color{lightred}-4.8})&67.9 ({\color{lightred}-3.0})\\
    GLTR&62.6&47.2 ({\color{darkred}-15.4})&31.2 ({\color{darkred}-31.4})&59.8 ({\color{lightred}-2.8})&24.3 ({\color{darkred}-38.3})&45.8 ({\color{darkred}-16.8})&52.1 ({\color{darkred}-10.5})\\
    Binoculars&79.6&80.3 ({\color{cyan}+0.7})&43.5 ({\color{darkred}-36.1})&78.0 ({\color{lightred}-1.6})&37.7 ({\color{darkred}-41.9})&70.1 ({\color{lightred}-9.5})&74.3 ({\color{lightred}-5.3})\\
    GPTZero&66.5&64.0 ({\color{lightred}-2.5})&61.0 ({\color{lightred}-5.5})&65.1 ({\color{lightred}-1.4})&66.2 ({\color{lightred}-0.3})&66.2 ({\color{lightred}-0.3})&61.0 ({\color{lightred}-5.5})\\
    Originality&85.0&96.7 ({\color{blue}+11.7})&96.5 ({\color{blue}+11.5})&78.6 ({\color{lightred}-6.4})&9.3 ({\color{darkred}-75.7})&84.9 ({\color{lightred}-0.1})&71.4 ({\color{darkred}-13.6})\\
    \bottomrule
    \end{tabular}
    \caption{Accuracy Score at FPR=5\% for select detectors across different adversarial attacks. Colors indicate an {\color{blue}increase}, {\color{cyan}slight increase}, {\color{lightred}slight decrease}, and {\color{darkred}decrease} in performance. We see that not all adversarial attacks affect models equally---with some occasionally even improving performance of detectors instead of harming them.}
    \label{tab:adversarial_comparison}
\end{table*}

\paragraph{Finding 4: Seemingly strong, robust detectors can perform unexpectedly poorly}

The most accurate detectors---FastDetectGPT, Originality, Binoculars, etc.---may seem like reliable solutions to detection in general, but they sometimes deteriorate from perfect accuracy to complete failure (see Table \ref{tab:decoding_full_results}). The changes in experimental settings were not particularly sophisticated either: simply changing the text generator, switching decoding strategies, or applying a repetition penalty was enough to introduce up to 95+\% error rate. Our findings show that detectors tend not to generalize across different models or generation settings in the same domain. Compounded by the lack of evaluation at different false positive rates, domain-specific detectors for critical issues like fake news and education are particularly at risk of mislabeling human-written text as machine-generated without our awareness.

\paragraph{Finding 5: Detectors perform better on domains and models seen during training.}

Figure \ref{fig:heatmap} shows the performance of RoBERTa-Large GPT2, GPTZero, and RADAR on a cross-section of models and domains from RAID. We see that RoBERTa GPT2 achieves 95+\% accuracy on five domains generated by GPT2, but it rarely achieves beyond 60\% accuracy on text of the same domain from different models. This detector is open-source, so we know that it was trained exclusively on GPT2 in an open-domain setting. We observe similar trends with RADAR, as it performs uncharacteristically poorly when detecting movie reviews regardless of generative model.

All detectors known to have constrained training data skew heavily towards test data with similar characteristics, leading us to believe that detectors perform better on domains and models seen during training. Some closed-source models such as GPTZero display similar behavior, allowing us to infer what data was used to train them. These findings demonstrate the need for multi-generator training corpora, especially since many publicly available neural detectors focus on only one or two generative models \cite{guo2023close}.

\paragraph{Finding 6: Different detectors are vulnerable to different types of adversarial attacks}
In Table \ref{tab:adversarial_comparison}, we see that Binoculars and other metric-based methods degrade as much as 36.1\% when a small portion of words are swapped with synonyms. All detectors were sensitive to homoglyph attacks except for GPTZero which sustained only a 0.3\% loss under the homoglyph attack while five others dropped an average of 40.6\%. Detectors like RADAR that underwent adversarial training, unsurprisingly, were much more robust to adversarial attacks. 
These detector-dependent differences in vulnerability suggest that attacking an arbitrary detector without prior knowledge of the detector type or training distribution will be difficult. Adversaries may respond by attempting to discover what the detector was trained on---which our findings have shown could be possible---or attacking detectors with repeated queries.

In addition, we see that detector accuracy sometimes increases after an adversarial attack. RoBERTa GPT2, for example, improved after texts were paraphrased with T5 and after words were replaced with BERT-based synonyms. GPT2, RoBERTa, T5, and BERT are contemporaneous models trained on similar data, leading us to believe that detectors benefit from adversarial attacks that inadvertently modify text to be more similar to their training data. Our previous findings on the influence of training data on performance reinforce our hypothesis.

\section{Conclusion}
As the generation capabilities of language models have continued to increase, accurately and automatically detecting machine-generated text has become an important priority. Detection efforts have even surpassed the bounds of natural language processing research, spurring discussions by social media companies and governments on possibly mandating labels for machine-generated content. Despite the protective intentions of these mandates, our work shows that such regulations would be difficult to enforce even if they were implemented. Detectors are not yet robust enough for widespread deployment or high-stakes use: many detectors we tested are nearly inoperable at low false positive rates, fail to generalize to alternative decoding strategies or repetition penalties, show clear bias towards certain models and domains, and quickly degrade with simple black-box adversarial attacks.

The bulk of our findings may sound bleak, but we did uncover promising signs of improvement. Binoculars, for example, performed impressively well across models even at extremely low false positive rates, Originality achieved high precision in some constrained scenarios, and GPTZero was unusually robust to adversarial attacks. We believe that openly evaluating detectors on large, diverse, shared resources is critical to accelerating progress---and trust---in detection. Evaluating robustness is particularly important for detection, and it only increases in importance and the scale of public deployment grows. 

We also need to remember that detection is just one tool for a larger, even more valuable motivation: preventing harm by the mass distribution of text. Detecting machine-generated text was a useful proxy for identifying harmful text for a long time, but language models have improved to the point that generated text is frequently legitimate and not harmful \citep{schuster-etal-2020-limitations}. Therefore, detecting specific harmful elements---like misinformation, hate speech, and abuse---should take precedence over whether or not the text was authored by a machine. Knowing if a text was machine-generated, however, does still offer insights on the types of errors we can expect or the recency of the facts cited within. We hope that our analyses and the RAID dataset are a step toward a future in which AI detection tools are safely integrated into society as a multi-pronged approach to reducing harm. We encourage future work to build on this by including more models, languages, and generation settings in future shared resources.

\section*{Limitations}
While we attempt to cover a wide variety of domains, models, decoding strategies and adversarial attacks in our dataset, we recognize that there can never be a truly comprehensive dataset for robustness. In particular, our dataset lacks the inclusion of multilingual text in many diverse domains. We release our RAID-extra data to help begin this process but we acknowledge the limited nature of this approach (only having multilingual text in the news domain). We encourage future work to expand on our foundation and use our tools to create truly robust shared benchmarks in many languages.

Furthermore, as the state-of-the-art in language modeling continues to improve, datasets of generated text will naturally obsolesce and will need to be continually maintained with new generations. This creates issues with shared evaluations as detectors will need to be re-run on any new dataset items and any accuracy metrics will have to be updated. While we believe this dataset will continue to be useful for many years, we do acknowledge this limitation and plan to alleviate this by occasionally releasing new updated versions.

Finally, and most importantly, the concept of a public benchmark for out-of-domain robustness is an inherently limited one. As practitioners seek to improve performance on our benchmark they will undoubtedly specialize to the particular aspects of robustness we cover. This will lead to overfitting, even if detectors are not explicitly trained on examples. Such overfitting will result in the reappearance of exactly the problems we wished to alleviate by creating this dataset, namely that detector accuracies are generally over-reported. We trust that this process will, to some extent, be alleviated by regular releases of new versions and keeping a set of hidden test data private. That being said, it does not nullify the utility of the dataset as a resource for profiling robustness of classifiers.

\section*{Ethics Statement}
Detecting generated text is often accusatory in nature and can frequently result in disciplinary or punitive action taken against the accused party. This can cause significant harm even when detectors are correct, but especially when they are incorrect. This is especially problematic given recent work by \citet{liang2023gpt} showing that detectors are biased against non-native English writers. Our results also support this and suggest that the problem of false positives remains unsolved.

For this reason, we are opposed to the use of detectors in any sort of disciplinary or punitive context and it is our view that poorly calibrated detectors cause more harm than they solve. Therefore, until better evaluation standards are widely adopted in the detection community, the use of detectors in this fashion should be discouraged. We intend for our work to be the start of this conversation and look forward to a future where machine-generated detectors are deployed in safe and responsible ways. 

\section*{Acknowledgements}
The authors would like to thank the members of the lab of Chris Callison-Burch for their testing and detailed feedback on the contents of this paper. In addition, we'd like to thank Professor Dan Roth for his early and enthusiastic support of the project.

This research is supported in part by the Office of the Director of National Intelligence (ODNI), Intelligence Advanced Research Projects Activity (IARPA), via the HIATUS Program contract \#2022-22072200005. The views and conclusions contained herein are those of the authors and should not be interpreted as necessarily representing the official policies, either expressed or implied, of ODNI, IARPA, or the U.S. Government. The U.S. Government is authorized to reproduce and distribute reprints for governmental purposes notwithstanding any copyright annotation therein.

\bibliography{custom}

\appendix

\begin{figure*}
    \centering
    \includegraphics[width=2\columnwidth]{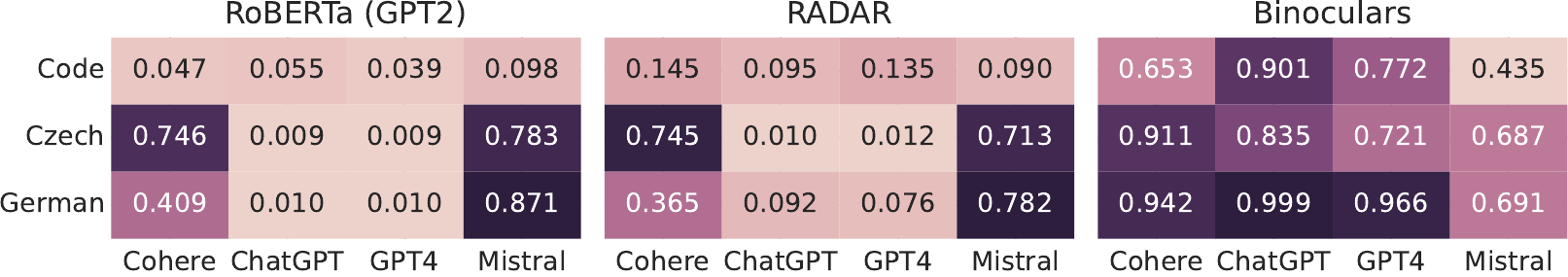}
    \caption{Heatmaps of accuracy for three of our detectors on German News, Python Code, and Czech News generations. We see that metric-based detectors have an edge over neural detectors in their ability to generalize to these unusual domains.}
    \label{fig:extra_domains_heatmap}
\end{figure*}

\section{Experiments on Multilingual and Code Generations (RAID-Extra)}
\label{app:multilingual}
In addition to the main RAID dataset we also release RAID-Extra: a collection of 2.3M generations in three extra challenging domains: Python Code, Czech News, and German News. These extra experiments were not included in the main benchmark as we felt that they were out of scope for most detectors and should not be used as a basis for comparison. Nonetheless, we were still curious to see what sorts of insights they can give us on detector performance.

\subsection{Data Generation}
Following \citet{macko-etal-2023-multitude}, multilingual prompts were written in the target language by a native speaker rather than being written in English and explicitly requesting that the model complete the generation in the target language. We found this to be the most effective method to get our generative models to adhere to the target language.

For Python Code generations, we applied an additional post-processing step as, in this domain, generative models had a tendency to write code between sets of triple backtick characters (\textasciigrave\textasciigrave\textasciigrave) and give natural language explanations of the code outside of the backticks. Thus for this domain and this domain only, we extracted the text between these sets of backticks and discarded all others. This was done to ensure that detectors could not use text descriptions of code for detection and would instead have to rely on the code itself.

\subsection{Results}
In Table \ref{tab:raid_extra_results} we report the accuracies of our 12 detectors on generations from RAID-extra at 5\% FPR. We see an interesting trend, that being the relatively strong performance of metric-based classifiers as compared to neural and commercial detectors. We suspect that metric-based classifiers are particularly well suited for such rare domains as they can be given any generative model to calculate their probabilities.

In Figure \ref{fig:extra_domains_heatmap} we show a heatmap of the performance of our detectors across the extra domains from select models. We see that Binoculars performs decently well when detecting Czech news articles despite the underlying generative model, Falcon 7B \cite{almazrouei2023falcon} being trained with five times as much German data as Czech data \cite{penedo2023refinedweb}. This seems to suggest that strong metric-based detectors for low-resource languages can be bootstrapped from highly-multilingual language models. Future work is necessary to understand the optimal setup in such scenarios.

\begin{table}[t]
    \small
    \centering
    \begin{tabular}{l|c|c|c|c}
    \toprule
    &\textbf{Code}&\textbf{Czech}&\textbf{German}&\textbf{Total}\\
    \midrule
    R-B GPT2&13.4&48.4&39.7&38.2\\
    R-L GPT2&12.7&53.1&48.4&43.5\\
    R-B CGPT&24.0&38.7&51.5&41.1\\
    RADAR&12.9&51.1&53.2&44.7\\
    \midrule
    GLTR&40.7&51.9&68.9&56.7\\
    F-DetectGPT&51.1&55.2&75.5&62.7\\
    LLMDet&17.5&24.0&10.6&17.3\\
    Binoculars&\textbf{59.9}&67.0&76.7&\textbf{69.6}\\
    \midrule
    GPTZero&33.8&33.6&49.5&39.0\\
    Originality&8.5&69.8&\textbf{89.1}&55.8\\
    Winston&24.5&\textbf{70.3}&73.8&56.2\\
    ZeroGPT&13.8&49.3&51.7&38.3\\
    \bottomrule
    \end{tabular}
    \caption{Accuracy of our 12 detectors at FPR=5\% on RAID-extra domains (Python Code, Czech News, and German News). We see that metric based detectors generally perform better than neural detectors.}
    \label{tab:raid_extra_results}
\end{table}

\begin{table*}[t]
    \small
    \centering
    \begin{tabular}{l|c|c|c|c|c|c|c|c|c|c}
    \toprule 
    &$\tau$&\textbf{News}&\textbf{Wiki}&\textbf{Reddit}&\textbf{Books}&\textbf{Abstracts}&\textbf{Reviews}&\textbf{Poetry}&\textbf{Recipes}&\textbf{Total}\\
    \midrule
    R-B GPT2&0.759&1.0\%&3.2\%&3.7\%&4.0\%&1.6\%&3.1\%&15.4\%&7.3\%&5.0\%\\
    R-L GPT2&0.307&1.8\%&7.0\%&2.4\%&2.5\%&1.3\%&4.0\%&15.4\%&5.1\%&5.0\%\\
    R-B C-GPT&0.988&4.4\%&4.0\%&0.7\%&9.0\%&0.0\%&1.1\%&0.5\%&18.6\%&5.0\%\\
    RADAR&0.343&0.2\%&1.2\%&10.7\%&2.0\%&4.2\%&20.4\%&8.6\%&0.1\%&5.0\%\\
    \midrule
    GLTR&0.818&0.4\%&0.8\%&1.1\%&0.0\%&0.2\%&0.1\%&1.8\%&33.4\%&5.0\%\\
    F-DetectGPT&0.920&5.1\%&1.6\%&1.9\%&6.3\%&1.8\%&2.8\%&7.5\%&2.2\%&3.7\%\\
    LLMDet&1.000&10.2\%&12.8\%&6.1\%&2.7\%&0.2\%&3.9\%&3.4\%&0.1\%&5.0\%\\
    Binoculars&0.090&2.8\%&5.7\%&7.2\%&4.0\%&5.3\%&5.7\%&3.8\%&5.9\%&5.0\%\\
    \midrule
    GPTZero&0.068&3.0\%&2.0\%&3.0\%&13.0\%&10.0\%&5.0\%&0.0\%&1.0\%&4.6\%\\
    Originality&0.494&4.0\%&13.0\%&2.0\%&3.0\%&4.0\%&4.0\%&3.0\%&7.0\%&5.0\%\\
    Winston&0.892&0.0\%&10.0\%&0.0\%&13.0\%&0.0\%&0.0\%&4.0\%&13.0\%&5.0\%\\
    ZeroGPT&1.000&29.0\%&48.0\%&0.0\%&1.0\%&0.0\%&5.0\%&0.0\%&52.0\%&16.9\%\\
    \bottomrule
    \end{tabular}
    \caption{False Positive Rates of our detectors on the RAID dataset broken up by domain when naively using a single threshold ($\tau$). We see that while the total FPR across all domains still sums to 5\%, individual domains see substantial fluctuation in FPR values.}
    \label{tab:naive_thresholds_fpr}
\end{table*}

\section{Fixed FPR Accuracy vs. F1 Score}
\label{app:accuracy_vs_f1}
Throughout our work we report accuracy on machine generated text at a set FPR because we believe it is the most intuitive way of understanding the performance of models in high-risk scenarios (i.e. “What percentage of generations are detected given that we tolerate an x\% chance of wrongly accusing someone”). Reporting the more standard F1 score is not only less intuitive but also treats false positives and false negatives as equivalent---which is not the case when dealing with high-risk scenarios or ones where detectors are repeatedly applied to texts from the same author.

In addition, since our dataset has a roughly 40:1 ratio of generated to human-written text, precision scores will artificially favor true positives over false positives as most all examples in the dataset are positive examples. However, in the real world, this ratio is reversed and the majority of texts are human written. Thus precision scores systematically over-represent the capabilities of detectors when used as a metric on a dataset like ours. We hope that our work can help to shed light on this issue and how easy it is to accidentally over-represent the performance of classifiers.

\section{Per-Domain Threshold Tuning}
\label{app:threshold_tuning}
When tuning the False Positive Rate of classifiers to a specific percentage (in our case 5\%), it is important to look not just at total FPR across all human texts, but also at FPR for each individual domain in the dataset. In our work, we ensure that classification thresholds are determined on a per-domain basis, i.e. that the FPRs of every detector on every domain of the data should be 5\%. While this undoubtedly adds complexity to the evaluation, it is an important step to ensure that detectors are being evaluated fairly with respect to one another (see Appendix \ref{app:threshold_search} for details about the threshold searching procedure). 

To drive home the importance of this point, in Table \ref{tab:naive_thresholds_fpr} we show the FPR of each classifier at a 5\% total FPR threshold broken up by domain. As we can see, while the total FPR is consistently 5\%, many detectors have particularly acute domain-specific weaknesses: RADAR has a 20.4\% FPR on Reviews, GLTR has a 33.4\% FPR on Recipes, and Originality has a 13\% FPR on Wikipedia. 

This asymmetric variation of FPR creates a dampening effect whereby the inclusion of weaker, more obscure domains reduces the accuracy of a classifier on more common domains---ultimately lowering total accuracy in the process. 

In order to avoid this issue, we ensure that our thresholds are chosen on a per-domain basis. That is, we find the threshold for each detector for each domain that results in 5\% FPR on that domain (see Table \ref{tab:thresholds}).

\section{Leaderboard and Pypi Package}
\label{app:leaderboard}
In order to truly achieve our goal of standardizing detector evaluation, it is important for RAID to not only be sufficiently challenging, but also have a simple, straightforward interface for submission and comparison. As discussed in section \ref{sec:release_structure}, we solve this problem by releasing a shared leaderboard and a Pypi package to make submitting to the leaderboard easy. The RAID package can be installed by running:
\begin{minted}{bash}
$ pip install raid-bench
\end{minted}
In Figure \ref{fig:raid_code} we show how to use our pypi package to load the RAID test set and run a detector on the texts. After getting the \texttt{predictions.json} file, submitting to the leaderboard simply involves making a folder in the repository, filling out metadata, and creating a pull request.

\begin{figure*}[t]
\frame{\includegraphics[width=\textwidth]{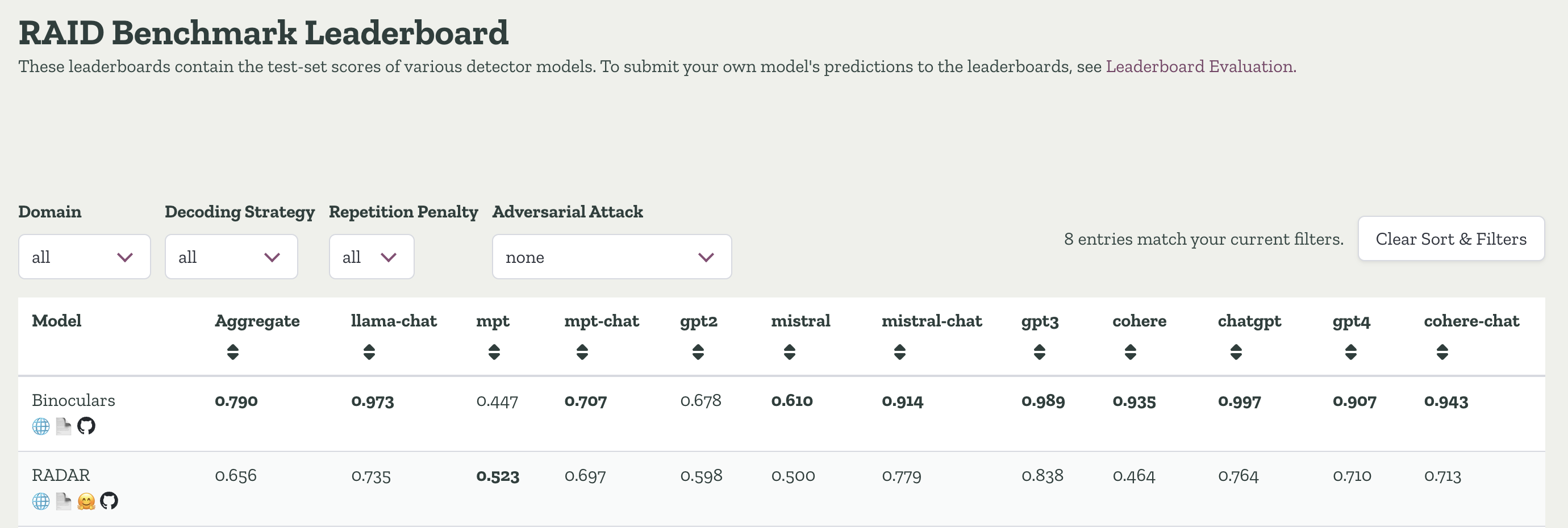}}
    \caption{Screenshot of the RAID leaderboard accessible at \url{https://raid-bench.xyz/leaderboard}}
    \label{fig:leaderboard}
\end{figure*}

\begin{figure}[t]
\begin{minted}{python}
from raid import run_detection
from raid.utils import load_data

# Define your detector function
def my_detector(texts: list[str]) -> list[float]:
    pass

# Load the RAID test data
test_df = load_data(split='test')

# Run your detector on the dataset
predictions = run_detection(my_detector, test_df)

# Write predictions to a JSON file
with open('predictions.json') as f:
    json.dump(predictions, f)
\end{minted}
\caption{A example showing how to use the RAID Pypi package to evaluate a detector on the dataset and submit it to the leaderboard.}
\label{fig:raid_code}
\end{figure}

In Figure \ref{fig:leaderboard} we show a screenshot of the leaderboard page on our project website. Submissions are open to the public and are automatically accepted and evaluated via pull requests to our git repository. The data used for evaluation on the leaderboard is the 10\% test split of the core RAID dataset that is released without labels.

\section{Dataset Details}
\label{app:dataset}
\subsection{Domains}
\label{app:domains}
In Table \ref{tab:human_sources} we report the exact number of documents sampled from each domain. The only two datasets that include post-2021 documents are Wikipedia and Abstracts. The Reviews domain did not have 2,000 documents and so we sampled the maximum amount. For each domain, we provide a detailed list of all human-written sources with metadata along with the RAID dataset in our code repository. A detailed description of the contents of each domain is as follows:

\paragraph{Book summaries} \cite{bamman2013new} This dataset contains plot-centric summaries of books along with their titles. We chose this dataset due to the first-person narrative style and because we expect generators and detectors with knowledge of the source material to have an advantage.

\paragraph{BBC News Articles} \cite{greene06icml} This dataset contains BBC articles with associated titles. The articles are spread out evenly across 5 categories (sport, technology, entertainment, politics, and business). This dataset was chosen since good generation requires factuality and because News is a large area for LLM-based harm.

\paragraph{Poems} \cite{poemsdataset} This dataset contains poems collected from \texttt{poemhunter.com} with their titles and genre. The poems are randomly spread out over genres and topics. We hypothesize that LLMs will write generic and repetitive poetry and that this tendency should be detectable. 

\paragraph{Recipes} \cite{bien-etal-2020-recipenlg} This dataset consists of recipes and their dish names. Recipes are a combination of a list of ingredients and a numbered list of steps. This dataset is difficult because it requires significant common sense reasoning, which is difficult for models.

\paragraph{Reddit Posts} \cite{volske-etal-2017-tl} This dataset contains reddit posts and their titles. We hypothesize that such data will be challenging to detect due to the first-person and informal style.

\paragraph{Movie Reviews} \cite{maas-etal-2011-learning} This dataset contains movie reviews from IMDb along with the names of the movies. The formality of the reviews are varied and this tests model's ability to recall details from movies as well as generate and detect opinionated text.

\paragraph{Wikipedia} \cite{aaditya_bhat_2023} This dataset contains introductions to various Wikipedia articles. This dataset is challenging as it tests the models ability to accurately recall facts relating to specific historical events.

\paragraph{Python Code} \cite{raychev-etal-2016-probabilistic} This dataset contains python solutions to coding problems and the associated problem title. We include this as an initial foray into the detection of AI-generated code.

\paragraph{Czech News} \cite{boháček2022finegrained} This domain consists of Czech language news articles. The topics and sources are diverse sampling from over 45 publications including mainstream journalistic websites, tabloids and independent news outlets in the Czech Republic.

\paragraph{German News} \cite{dietmar-etal-2017-german} This domain consists of German language news articles from DER STANDARD, an Austrian daily broadsheet newspaper. Articles are fairly political in nature covering topics such as the European migrant crisis, the 2016 Austrian presidential elections and the Syrian Civil War. We expect this dataset to test models' ability to generate opinionated political content in another language.

\paragraph{Paper Abstracts} \cite{arxivdataset} This is a dataset of abstracts scraped from ArXiv together with paper titles. For this dataset and this dataset only, we filter the data such that only papers from 2023 or later are present in the data. This allows us to rule out the possibility that our models have memorized this text.

\begin{table}[t]
\centering 
\small
\begin{tabular}{l|c|c} 
\toprule
\textbf{Dataset}&\textbf{Genre}&\textbf{Size}\\
\midrule
\cite{arxivdataset}&Abstracts&1966\\
\cite{bamman2013new}&Books&1981\\
\cite{raychev-etal-2016-probabilistic}&Code&920\\
\cite{greene06icml}&News&1980\\
\cite{poemsdataset}&Poetry&1971\\
\cite{bien-etal-2020-recipenlg}&Recipes&1972\\
\cite{volske-etal-2017-tl}&Reddit&1979\\
\cite{maas-etal-2011-learning}&Reviews&1143\\
\cite{aaditya_bhat_2023}&Wiki&1979\\
\cite{boháček2022finegrained}&Czech&1965\\
\cite{dietmar-etal-2017-german}&German&1970\\
\bottomrule
\end{tabular}
\caption{The number of articles sampled from each domain with their corresponding sources}
\label{tab:human_sources}
\end{table}

\subsection{Generative Models}
\label{app:models}
In Table \ref{tab:generative_models} we list the exact generative models used in our project along with their unique identifiers. All open-source models were run using the HuggingFace \texttt{transformers} library \cite{wolf2020huggingfaces} and all closed-source models were run using the proprietary APIs from Cohere\footnote{\url{https://docs.cohere.com/reference/about}} and OpenAI\footnote{\url{https://platform.openai.com/docs/introduction}}. The following is a detailed list of the generative models used in the project.

\begin{table}
\centering 
\small
\begin{tabular}{l|l} 
\toprule
\textbf{Model}&\textbf{Identifier}\\
\midrule
\textbf{GPT-2}&\texttt{gpt2-xl}\\
\cite{radford2019language}&\\
&\\
\textbf{MPT} (\textit{+ Chat}) &\texttt{mpt-30b}\\
\cite{MosaicML2023Introducing}&\texttt{mpt-30b-chat}\\
&\\
\textbf{Mistral} (\textit{+ Chat})&\texttt{Mistral-7B-v0.1}\\
\cite{jiang2023mistral}&\texttt{Mistral-7B-Instruct-v0.1}\\
&\\
\textbf{LLaMA Chat}&\texttt{Llama-2-70b-chat-hf}\\
\cite{touvron2023llama}&\\
&\\
\textbf{Cohere} (\textit{+ Chat})&\texttt{command (co.generate())}\\
\cite{Cohere2024}&\texttt{command (co.chat())}\\
&\\
\textbf{GPT-3}&\texttt{text-davinci-002}\\
\cite{ouyang2022training}&\\
&\\
\textbf{ChatGPT}&\texttt{gpt-3.5-turbo-0613}\\
\cite{chatgpt-etal-2022}&\\
&\\
\textbf{GPT-4}&\texttt{gpt-4-0613}\\
\cite{openai2023gpt4}&\\
\bottomrule
\end{tabular}
\caption{The generative models used in our project}
\label{tab:generative_models}
\end{table}

\paragraph{GPT2 XL 1.5B}\cite{radford2019language} is a decoder-only model trained on the WebText dataset. This dataset consists of a collection of all documents that were linked from reddit posts or comments that had at least 3 or more upvotes. Released in February of 2019 and having 1.5B parameters, GPT2 is the predecessor of GPT3 and GPT4 and the most powerful open-source OpenAI model. 

\paragraph{GPT3}\cite{ouyang2022training} is a closed-source language model released by OpenAI on November 29th, 2022. The model was allegedly trained with a variety of data including the Common Crawl (filtered), WebText2, and Wikipedia datasets \cite{brownetal2020} but exact composition of the training dataset is unknown. It is the first model shown to work well with prompts and has shown great zero- and few-shot capabilities. In this study, we use the \texttt{text-davinci-002} model. We queried the model from November 1st to November 2nd 2023. Unfortunately, as of January 4th 2024, this model is no longer available for use on the OpenAI API. This is unfortunate as it prevents us from expanding the domains in future releases. We encourage researchers to keep this in mind when using OpenAI models for their research projects.

\paragraph{ChatGPT}\cite{chatgpt-etal-2022} is a version of GPT3 fine-tuned using Reinforcement Learning from Human Feedback (RLHF) \cite{ouyang2022training}. We use the June 13th 2023 checkpoint of the model (\texttt{gpt-3.5-turbo-0613}). Although the number of parameters is unknown, ChatGPT demonstrates outstanding capability in language and code generation. 

\paragraph{GPT4}\cite{openai2023gpt4} is the latest iteration of OpenAI's GPT family of models and is one of the largest and most powerful language models available to date. In this study, we use the \texttt{gpt-4-0613} checkpoint of the model through the \texttt{ChatCompletion} interface\footnote{\url{https://platform.openai.com/docs/guides/chat}}. We queried the model from November 1st to November 2nd 2023.

\paragraph{LLaMA 2 70B}\cite{touvron2023llama} is a decoder-only model trained by Meta (Facebook) and is the second model in the LLaMA series. Released on July 18th 2023, LLaMA 2 is the successor to the original LLaMA model which was trained on webpages from CommonCrawl, multilingual Wikipedia, books from Project Gutenberg, and QAs from Stack Exchange. The composition of LLaMA 2's training data is not known but it has shown impressive performance on many open-source evaluations and is widely considered competitive with the open-source state-of-the-art.

\paragraph{Mistral 7B}\cite{jiang2023mistral} is a decoder-only model trained by Mistral and is the first model released by the company. Released on September 27, 2023, Mistral 7B outperforms LLaMA 2 13B across various benchmarks at half the size. While model weights are open-source, the training data for the model is not and no details have been released regarding the makeup of this data.

\paragraph{MPT 30B}\cite{MosaicML2023Introducing} is a decoder-only model trained by Mosaic and is the first model released by the company. Released on June 22nd 2023, MPT 30B has an 8k context window and outperforms GPT-3 on various reasoning tasks. Training data consists of deduplicated C4 \cite{raffel2020exploring,lee-etal-2022-deduplicating}, the RedPajama\footnote{\url{https://www.together.ai/blog/redpajama-data-v2}} split of CommonCrawl, and selected programming languages from The Stack \cite{Kocetkov2022TheStack}.

\paragraph{Cohere \textit{command}}\cite{Cohere2024} is a closed-source model trained and released by Cohere. While original versions of this model were quoted as having roughly 50 billion parameters \cite{liang2023holistic}, the size and training data of the current version of the Cohere model is unknown. Unlike OpenAI, Cohere does not version the \textit{command} model and so, much like with \texttt{text-davinci-002} we are unable to expand our dataset to new domains without re-generating all generations. We queried the command model through both the \texttt{co.generate()} and \texttt{co.chat()} endpoints from November 1st to November 2nd 2023.

\subsection{Prompts}
\label{app:prompts}
In Table \ref{tab:generation_prompts} we report the prompt templates for each of the 8 domains used in our project. We specifically avoided biasing the model towards a particular length or style of generation and did not use any of the text from the human-written documents other than the title when prompting. 

To decide on the particular form of our prompts, we conducted two rounds of manual review. These rounds consisted of the authors determining problematic prompts by manually reviewing 10 generations per model in each domain for instances of degenerate repetition, meta-commentary or other signs of generated output. In problematic domains, we conducted individual explorations to determine if there existed some high level prompting concept that removed the unintended behavior and whether or not using such a prompt style caused regressions across other generators. After identifying a desired prompting change, we re-wrote our templates and restarted the review process.

Through this investigation, we found that continuation-style models benefit greatly from explicitly stating the source website in the prompt and that chat-style models benefit from language explicitly asking them not to repeat the title of the article (See Table \ref{tab:generation_prompts}). While we were unable to remove all instances of degenerate repetition or meta-commentary, this investigation significantly increased the quality of our dataset and we encourage future work on dataset creation to conduct a similar process when engineering their prompts.

\begin{table}[t]
\centering 
\small
\begin{tabular}{p{0.35\linewidth}|c|p{0.4\linewidth}} 
\toprule
\textbf{Attack}&\textbf{$\theta$}&\textbf{Source}\\
\midrule
\textbf{Alternative Spelling}&100\%&\cite{liang2023gpt}\\
\textbf{Article Deletion}&50\%&\cite{liang2023mutationbased,guerrero2022mutationbased}\\
\textbf{Homoglyph}&100\%&\cite{wolff2020attacking,gagiano-etal-2021-robustness}\\
\textbf{Insert Paragraphs}&50\%&\cite{bhat-parthasarathy-2020-effectively}\\
\textbf{Number Swap}&50\%&\cite{bhat-parthasarathy-2020-effectively}\\
\textbf{Paraphrase}&100\%&\cite{krishna2023paraphrasing,sadasivan2023aigenerated}\\
\textbf{Misspelling}&20\%&\cite{liang2023mutationbased,gagiano-etal-2021-robustness,gao2018blackbox}\\
\textbf{Synonym}&50\%&\cite{pu-etal-2023-deepfake}\\
\textbf{Upper Lower}&5\%&\cite{gagiano-etal-2021-robustness}\\
\textbf{Whitespace}&20\%&\cite{cai2023evade,gagiano-etal-2021-robustness}\\
\textbf{Zero-Width Space}&100\%&\cite{guerrero2022mutationbased}\\
\bottomrule
\end{tabular}
\caption{The adversarial attacks used in the project. $\theta$ represents the manually determined fraction of available attacks carried out. We determine this fraction through manual review.}
\label{tab:adversarial}
\end{table}

\begin{table*}[t]
\centering 
\small
\begin{tabular}{l|p{0.41\linewidth}|p{0.41\linewidth}} 
\toprule
&\textbf{Continuation-Style Prompt}&\textbf{Chat-Style Prompt}\\
\midrule
\textbf{Abstracts}&The following is the full text of the abstract for a research paper titled "\{title\}" from arxiv.org:&Write the abstract for the academic paper titled "\{title\}".\\
\midrule
\textbf{Books}&The following is the full text of a plot summary for a novel titled "\{title\}" from wikipedia.org:&Write the body of a plot summary for a novel titled "\{title\}". Do not give it a title.\\
\midrule
\textbf{Code}&The following is the full text of a Python code solution to the exercise "\{title\}" from stackoverflow.com:&Write just the Python code solution for the problem "\{title\}".\\
\midrule
\textbf{German}&Schreiben Sie einen Nachrichtenartikel mit dem Titel "\{title\}".&Es folgt ein Nachrichtenartikel mit dem Titel "\{title\}":\\
\midrule
\textbf{Czech}&Následuje článek s názvem "\{title\}":&Napiš text článku, který má nadpis "\{title\}".\\
\midrule
\textbf{News}&The following is the full text of a news article titled "\{title\}" from bbc.com:&Write the body of a BBC news article titled "\{title\}". Do not repeat the title.\\
\midrule
\textbf{Poetry}&The following is the full text of a poem titled "\{title\}" from poemhunter.com:&Write the body of a poem titled "\{title\}". Do not repeat the title.\\
\midrule
\textbf{Recipes}&The following is the full text of a recipe for a dish called "\{title\}" from allrecipes.com:&Write a recipe for "\{title\}".\\
\midrule
\textbf{Reddit}&The following is the full text of a post titled "\{title\}" from reddit.com:&Write just the body of a Reddit post titled "\{title\}". Do not repeat the title.\\
\midrule
\textbf{Reviews}&The following is the full text of a review for the movie "\{title\}" from IMDb.com:&Write the body of an IMDb review for the movie "\{title\}". Do not give it a title.\\
\midrule
\textbf{Wiki}&The following is the full text of an article titled "\{title\}" from wikipedia.com:&Write the body of a Wikipedia article titled "\{title\}".\\
\bottomrule
\end{tabular}
\caption{The text of the generation prompts for all ten datasets in both continuation and chat style. The field \{title\} was replaced with the title of the book, dish, or news article before being passed into the generative model.}
\label{tab:generation_prompts}
\end{table*}

\subsection{Adversarial Attacks}
\label{app:adversarial}
In Table \ref{tab:adversarial} we list the attacks used along with the attack rate $\theta$ and the relevant sources for the attacks. In this section we will list the attacks in more detail and discuss the various design decisions made. Implementations for all attacks can be found in our GitHub repository.

\paragraph{Alternative Spelling} We use an American to British English dictionary\footnote{\url{https://github.com/hyperreality/American-British-English-Translator}} to construct a mapping between American and British spellings of words. We then find all instances of such words in the generation (defaulting to the longest available match if there were multiple substring matches for the same token). We then randomly sample a fixed percentage $\theta$ of the possible mutations to make with a set seed and apply the attack at those indices.

\paragraph{Article Deletion} We search through the text and find every instance of the articles ``a'', ``an'', and ``the''. We then randomly sample a fixed percentage $\theta$ of the possible mutations to make with a set seed and apply the attack at those indices.

\paragraph{Homoglyph} Homoglyphs are character that are non-standard unicode characters that strongly resemble standard English letters. These are typically characters used in Cyrillic scripts. We use the set of homoglyphs from \citet{wolff2020attacking} which includes substitutions for the following standard ASCII characters: a, A, B, e, E, c, p, K, O, P, M, H, T, X, C, y, o, x, I, i, N, and Z. We limit ourselves to only homoglyphs that are undetectable to the untrained human eye, thus we are able to use an attack rate of $\theta=100\%$ and apply the attack on every possible character. For characters that have multiple possible homoglyphs we randomly choose between the homoglyphs.

\paragraph{Insert Paragraphs} For this attack, we again split sentences using Punkt \cite{kiss-strunk-2006-unsupervised} and construct a list of all inter-sentence spans. We then sample $\theta$ percent of those spans and add the double newline character \texttt{\textbackslash n\textbackslash n} in-between the sentences to simulate a paragraph break.

\paragraph{Number Swap} For this attack we use the following regular expression to extract all instances of numerical digits in the generation: \texttt{"\textbackslash d+.?\textbackslash d*"}. We then randomly select $\theta$ percent of these digits to modify and for each digit we randomly select an alternate number between 0 and 9 to replace the character with.

\paragraph{Paraphrase} For paraphrasing we run the DIPPER-11B model\footnote{\url{https://huggingface.co/kalpeshk2011/dipper-paraphraser-xxl}} from \citet{krishna2023paraphrasing} through HuggingFace \cite{wolf2020huggingfaces}. DIPPER is a fine-tuned version of T5-11B \cite{raffel2020exploring} specifically made for paraphrasing text to avoid machine-generated text detectors. We use the default settings from the paper, namely a sentence interval of 3 with lexical diversity of 60 and order diversity of 0. Since paraphrases are not inherently noticeable when models are correct, we are able to apply this attack across the entirety of the output text ($\theta=100\%$).

\paragraph{Misspelling} For this attack, we manually constructed a dictionary of common misspellings\footnote{\url{https://en.wikipedia.org/wiki/Commonly_misspelled_English_words}} and only applied the attack to instances of words that have misspellings in our dictionary. We did this to minimize suspicion from human readers, as certain common words such as `the' or 'him' are rarely misspelled. Instead of randomly selecting $\theta$ percent of the possible candidate words to misspell, we follow \citet{gagiano-etal-2021-robustness} and misspell only the top $\theta$ percent most likely candidate words by log likelihood as determined by GPT 2 small. This allows us to choose only the most effective misspellings to apply.

\paragraph{Synonym} For this attack we originally planned to use the DFTFooler algorithm from \citet{pu-etal-2023-deepfake}. However, we found the candidate synonyms to be of low quality and easily detectable by our manual analysis. Thus we decided to implement our own algorithm based largely on DFTFooler. Our algorithm produces high-quality and diverse synonym substitutions without relying on any of the large decoder-only language models used for generation.

We start by iterating over all tokens in the generation. For each token $i$ we replace it with a mask token and get the top 20 most likely mask-fill candidates\footnote{In order to reduce computation time we limit BERT to a window of 20 tokens on either side of the index when determining the mask-fill candidates. We found no reduction in candidate quality from this modification.} according to BERT \cite{devlin-etal-2019-bert}. We then compute the part-of-speech tag for each candidate using NLTK \cite{bird-loper-2004-nltk} and reject all candidates that do not match the part-of-speech of the original token. We then get the static FastText \cite{bojanowski-etal-2017-enriching} embeddings for all candidate substitutions and reject all tokens that have cosine similarity of less than 0.5 with the original token. Doing this for each index $i$ gives us a global list of all valid candidate swaps across the entire generated passage. From this list we select the $\theta \cdot L$ most likely synonym swaps according to BERT where $L$ is the length (in tokens) of the passage. 

The full code for this algorithm can be found in our project repository along with the implementations of the other adversarial attacks.

\paragraph{Upper-Lower} This attack randomly selects some $\theta$ percent of the tokens in the passage and swaps the first letter of the token to be uppercase if it was lowercase and lowercase if it was originally uppercase.

\paragraph{Whitespace} This attack randomly selects some $\theta$ percent of inter-token spaces and adds an extra space character inbetween the tokens. This can occasionally result in multiple spaces added between two tokens as sampling is done with replacement.

\paragraph{Zero-Width Space} The unicode zero-width space \texttt{U+200B} is a character that exists in text encoding but is not visible to human readers in most scenarios. Thus, for this attack we insert this character at every possible opportunity (before and after each visible character in the generation).

\subsection{Repetition Penalty vs. Frequency Penalty vs. Presence Penalty}
\label{app:repetition_penalty}
Given a temperature $T > 0$ and a set of scores $x_i \in \mathbb{R}^d$ for each token $i$ in a vocabulary, the probability $p_i$ of predicting the $i$th token is given by:
$$p_i = \frac{\text{exp}(x_i / T)}{\sum_j\text{exp}(x_j / T)}$$

The repetition penalty defined by \citet{keskar2019ctrl} modifies this distribution as follows:
$$p_i = \frac{\text{exp}(x_i / (T \cdot I(i\in g)))}{\sum_j\text{exp}(x_j / (T \cdot I(j\in g)))}$$
Where $g$ is a list of previously generated tokens and $I(c) = \theta \text{ if c is True else 1}$. OpenAI implements a form\footnote{\url{https://platform.openai.com/docs/guides/text-generation/parameter-details}} of this penalty (referred to as a `presence penalty') which is additive instead of multiplicative: 
$$p_i = \frac{\text{exp}((x_i / T) - I(i\in g))}{\sum_j\text{exp}((x_j / T) - I(j\in g))}$$
Cohere (as of May 13th 2024) provides no documentation on the nature of their presence penalty despite requests from the authors for the proper documentation.

\subsection{Hardware}
We ran our generations over the course of 15 days from November 1st 2023 to November 15th 2023 on 32 NVIDIA 48GB A6000 GPUs. We ran all models with 16-bit precision as we found that outputs were identical to full precision and it cut our inference time in half. The amount of GPU hours used by each family of models is as follows: LLaMA 2 70B (+ Chat) 8,376 hours, MPT (+ Chat) 5,440 hours, Mistral (+ Chat) 672 hours, GPT2 (+ Chat) 352 hours. In total we used 14,872 GPU hours (620 GPU days) to generate the RAID dataset.

\section{Detector Details}
\label{app:detectors}

\subsection{Detectors}
In this section we provide a detailed description of all detectors used in the evaluations of the RAID dataset.

\paragraph{RoBERTa (GPT2)} \cite{solaiman-etal-2019} This detector\footnote{We download the model hosted by OpenAI from the following link \url{https://openaipublic.azureedge.net/gpt-2/detector-models/v1/detector-large.pt}} is a RoBERTa model \cite{liu2019roberta} fine-tuned on the GPT2 output dataset. This dataset consists of outputs from GPT2 in open domain settings with three different decoding strategies: greedy decoding, top-k=50, and fully random sampling and has been a baseline inclusion for many years. We use both the base and large size of this model in our comparisons.

\paragraph{RoBERTa (ChatGPT)} \cite{guo2023close} This detector is a RoBERTa-base model \cite{liu2019roberta} fine-tuned on the HC3 dataset. HC3 consists of roughly 27,000 questions paired with both human and ChatGPT answers in various domains such as reddit, medicine, finance, and law. We download and query the detector via HuggingFace datasets with the unique identifier \texttt{Hello-SimpleAI/chatgpt-detector-roberta}

\paragraph{RADAR} \cite{hu2023radar} This detector is a fine-tuned version of Vicuna 7B (which itself is a fine-tune of LLaMA 7B). It was trained in a generative adversarial setting alongside a paraphrase model. The paraphraser was trained specifically to fool the detector and the detector was trained to accurately detect generations from the paraphraser, human-text from the WebText dataset, and outputs from the original language model. We download and query this detector from HuggingFace with the unique identifier \texttt{TrustSafeAI/RADAR-Vicuna-7B}

\paragraph{GLTR} \cite{gehrmann-etal-2019-gltr} Originally intended as an interface to help humans better detect generated text, GLTR has become a standard baseline in robustness studies of detector abilities. GLTR evaluates the likelihood of text according to a language model and bins tokens according to their likelihoods and uses these bins as features for detection. We use the default settings from the GLTR repository\footnote{\url{https://github.com/HendrikStrobelt/detecting-fake-text}} namely our cutoff set at rank=10 and the language model set to GPT2 small.

\paragraph{FastDetectGPT} \cite{bao2023fastdetectgpt} This detector is an improvement on the original DetectGPT \cite{mitchell-etal-2023-detectgpt}---speeding up inference by 340x without any reduction in accuracy. For the scoring model we use the repository default of GPT-Neo-2.7B and for the reference model we again use the default of GPT-J-7B. Since neither of these models were used to generate continuations in our dataset, we felt that this was a reasonable choice.

\paragraph{Binoculars} \cite{hans2024spotting} This detector uses perplexity divided by cross-entropy between two very similar language models as the metric for detection. In our implementation we use the code from the official GitHub repository and calculate perplexity using the default models from the repository, namely Falcon 7B and Falcon 7B Instruct \cite{almazrouei2023falcon}. Much like FastDetectGPT, since neither of these models were used to generate continuations, we believed this made for a fair comparison.

\paragraph{LLMDet} \cite{wu-etal-2023-llmdet} This detector computes the proxy-perplexity of the input text from 10 different small language models and uses these features for detection. The proxy-perplexity is an approximation of the true perplexity calculated by repeatedly sampling n-grams from models rather than by actually running them. Of the models used, none of them were used for generations in our dataset, thus this was a fair comparison.

\paragraph{GPTZero} \cite{tian2023gptzero} GPTZero is a closed-source commercial detector released in January 2023 and was among the first to gain widespread media attention for their detection-as-a-service business model. We queried the detector on January 24th 2024 using the v2 API\footnote{\url{https://api.gptzero.me/v2/predict/text}} and threshold on the \texttt{completely\_generated\_prob} field.

\paragraph{Originality} Originality is a closed-source commercial detector released in November 2022 and was the first company to adopt the detection-as-a-service business model. We queried this detector from January 24th to 25th through the v1 API\footnote{\url{https://api.originality.ai/api/v1/scan/ai}} and threshold on the `\texttt{score}' field in the output JSON. For the Czech and German news domains, we specifically query the multilingual ``version 3'' of the detector.

\paragraph{Winston} This detector is the closed-source commercial detector that claims the highest accuracy out of any detector (99.98\%). We query this model through the v1 API\footnote{\url{https://api.gowinston.ai/functions/v1/predict}} and specifically set the input language to German when we detect that our input text is in German. Unfortunately, since winston does not support detection in Czech, we use English as the input language for the Czech news domain. We once again 
threshold on the `\texttt{score}' field in the output JSON.

\paragraph{ZeroGPT} This detector is the final commercial detector. We run this as it is the only detector besides GPTZero that has been evaluated in another benchmark paper. We query the API at the following link\footnote{\url{https://api.zerogpt.com/api/detect/detectText}} and use the `\texttt{isHuman}' return field as the classifier output.

\begin{table*}[t]
    \small
    \centering
    \begin{tabular}{l|c|c|c|c|c|c|c|c}
    \toprule 
    &\textbf{News}&\textbf{Wiki}&\textbf{Reddit}&\textbf{Books}&\textbf{Abstracts}&\textbf{Reviews}&\textbf{Poetry}&\textbf{Recipes}\\
    \midrule
    RoBERTa-B GPT2&0.032&0.379&0.477&0.586&0.055&0.539&0.998&0.916\\
    &(5.0\%)&(5.0\%)&(5.0\%)&(5.0\%)&(5.0\%)&(5.0\%)&(5.0\%)&(5.0\%)\\
    RoBERTa-L GPT2&0.070&0.459&0.100&0.161&0.085&0.298&0.762&0.315\\
    &(5.0\%)&(5.0\%)&(5.0\%)&(5.0\%)&(5.0\%)&(5.1\%)&(5.0\%)&(5.0\%)\\
    RoBERTa-B C-GPT&0.987&0.983&0.219&0.996&0.007&0.371&0.295&0.998\\
    &(5.0\%)&(5.0\%)&(5.0\%)&(5.0\%)&(5.0\%)&(5.0\%)&(5.0\%)&(5.0\%)\\
    RADAR&0.022&0.061&0.695&0.174&0.31&0.997&0.457&0.016\\
    &(5.0\%)&(5.0\%)&(5.0\%)&(5.0\%)&(5.0\%)&(5.0\%)&(5.0\%)&(5.0\%)\\
    \midrule
    GLTR&0.788&0.788&0.767&0.742&0.726&0.757&0.756&0.863\\
    &(5.0\%)&(5.0\%)&(5.0\%)&(5.0\%)&(5.0\%)&(5.0\%)&(5.0\%)&(5.0\%)\\
    FastDetectGPT&0.920&0.870&0.870&0.930&0.860&0.900&0.940&0.880\\
    &(4.8\%)&(5.1\%)&(5.1\%)&(4.9\%)&(4.6\%)&(4.8\%)&(5.9\%)&(5.5\%)\\
    LLMDet&1.000&1.000&1.000&1.000&0.999&1.000&1.000&0.998\\
    &(5.0\%)&(5.0\%)&(5.0\%)&(5.0\%)&(5.0\%)&(5.1\%)&(5.0\%)&(5.0\%)\\
    Binoculars&0.077&0.093&0.099&0.085&0.092&0.097&0.084&0.094\\
    &(4.9\%)&(5.0\%)&(5.0\%)&(5.0\%)&(5.0\%)&(4.9\%)&(5.0\%)&(5.0\%)\\
    \midrule
    GPTZero&0.047&0.032&0.057&0.125&0.125&0.070&0.031&0.035\\
    &(5.0\%)&(5.0\%)&(5.0\%)&(5.0\%)&(5.0\%)&(5.0\%)&(5.0\%)&(5.0\%)\\
    Originality&0.375&0.938&0.250&0.312&0.257&0.461&0.047&0.750\\
    &(5.0\%)&(5.0\%)&(5.0\%)&(5.0\%)&(5.0\%)&(5.0\%)&(5.0\%)&(5.0\%)\\
    Winston&0.001&0.970&0.062&0.998&0.000&0.062&0.875&0.996\\
    &(4.0\%)&(5.0\%)&(5.0\%)&(5.0\%)&(6.0\%)&(5.0\%)&(5.0\%)&(5.0\%)\\
    ZeroGPT(*)&1.000&1.000&0.375&0.250&0.500&1.000&0.125&1.000\\
    &(29.0\%)&(48.0\%)&(1.0\%)&(9.0\%)&(4.0\%)&(5.0\%)&(5.0\%)&(52.0\%)\\
    \bottomrule
    \end{tabular}
    \caption{Thresholds found by our search and the exact False Positive Rates on our dataset. We see that ZeroGPT is incapable of achieving the target FPR of 5\% in many domains.}
    \label{tab:thresholds}
\end{table*}

\subsection{Thresholds}
\label{app:threshold_search}
In order to find the thresholds that achieve a fixed false positive rates we had to implement some basic search procedure. We searched our thresholds in a linear fashion: We start at the threshold corresponding to the mean score of human data (50\% FPR) and approach the desired false positive rate by iteratively incrementing or decrementing the threshold. If we overshoot the target fpr we divide our step size in half and flip the sign. We continue to do this until the false positive rate is within $\epsilon=0.0005$ of the desired false positive rate or until 50 iterations are reached. If 50 iterations are reached without convergence then we select the threshold corresponding to the FPR that is closest to the target while still being less than the target. If there is no such threshold then we note that the detector could not reach the target FPR and simply choose the value closest to and greater than the target. In Table \ref{tab:thresholds} we list the thresholds our algorithm found for each detector along with the exact false positive rates that these thresholds allow.

\section{Extended Figures and Tables}
\label{app:extended-detail}

\subsection{Dataset Statistics and Evaluations}
\begin{figure*}
    \centering
    \includegraphics[width=2\columnwidth]{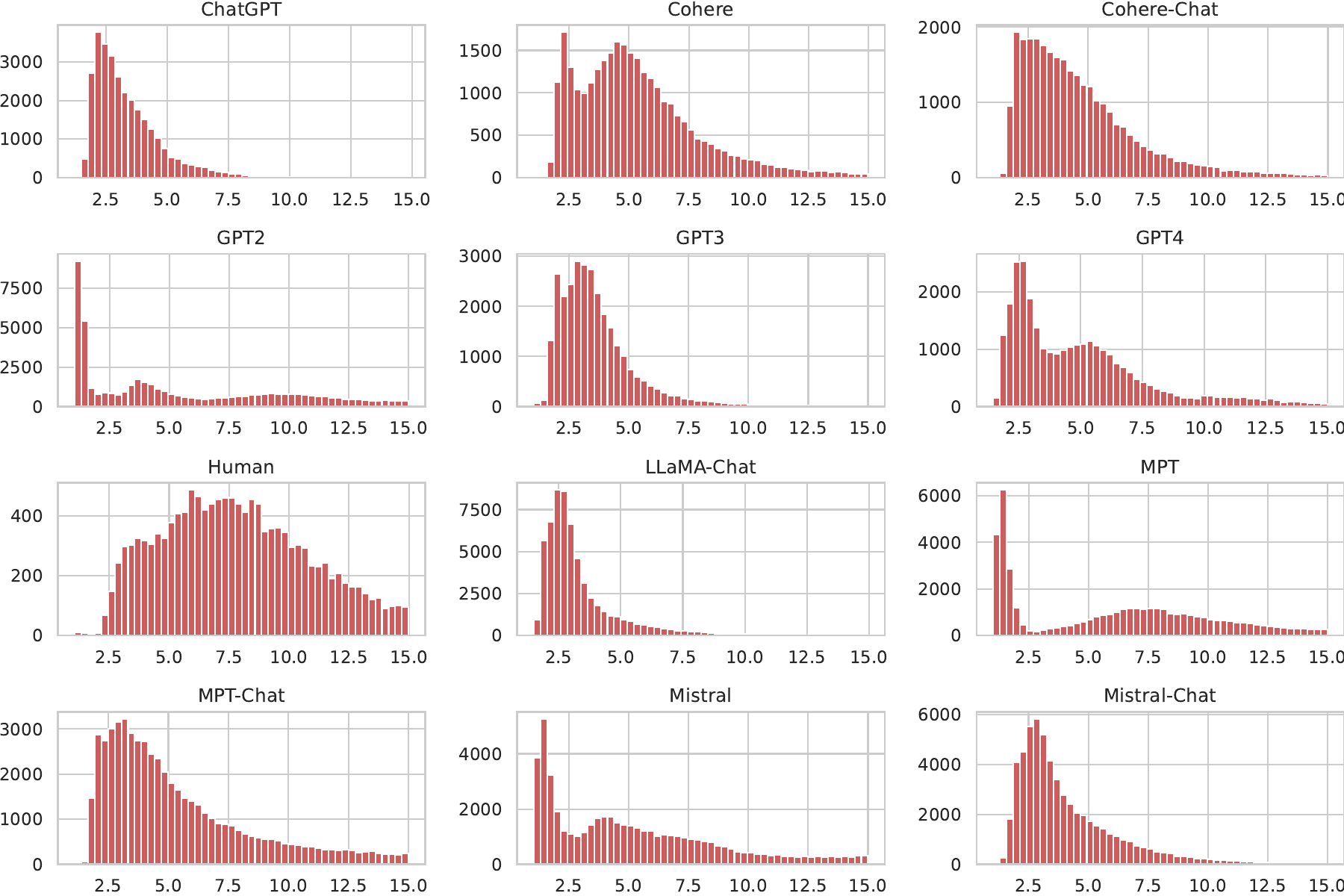}
    \caption{Histograms of perplexity according to LLaMA 2 7B per generative model in the dataset. We see that there is still a significant difference between human-written and machine-generated text with respect to perplexity.}
    \label{fig:app_ppl_llama2_permodel}
\end{figure*}

\begin{figure*}
    \centering
    \includegraphics[width=2\columnwidth]{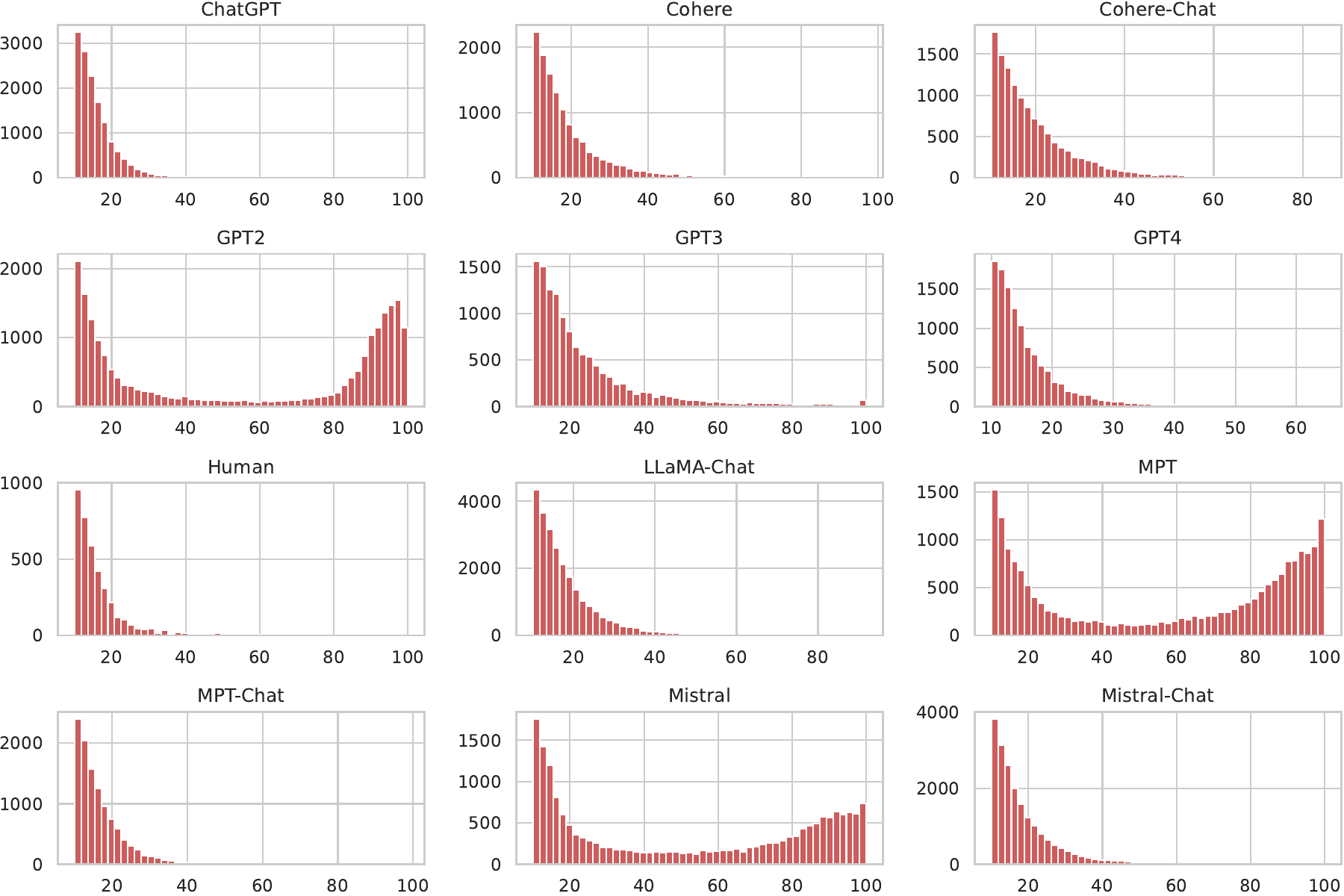}
    \caption{Histograms of SelfBLEU \cite{zhu2018texygen} per generative model in the dataset. We see that continuation models tend to be more repetitive than chat models and that human-written text is by far the least repetitive.}
    \label{fig:app_ppl_selfbleu_permodel}
\end{figure*}

In Figures \ref{fig:app_ppl_llama2_permodel} and \ref{fig:app_ppl_selfbleu_permodel} we report extended statistics about the dataset. We see some interesting trends here, namely that human-written text is still significantly less likely than machine-generated text according to LLaMA 2 7B and that it is also the least repetitive. These results push back on claims that language models are approaching human-level performance and therefore detection is unreasonably difficult.

\subsection{Extended Heatmaps}
\label{app:more-heatmaps}
\begin{figure*}
    \centering
    \includegraphics[width=\textwidth]{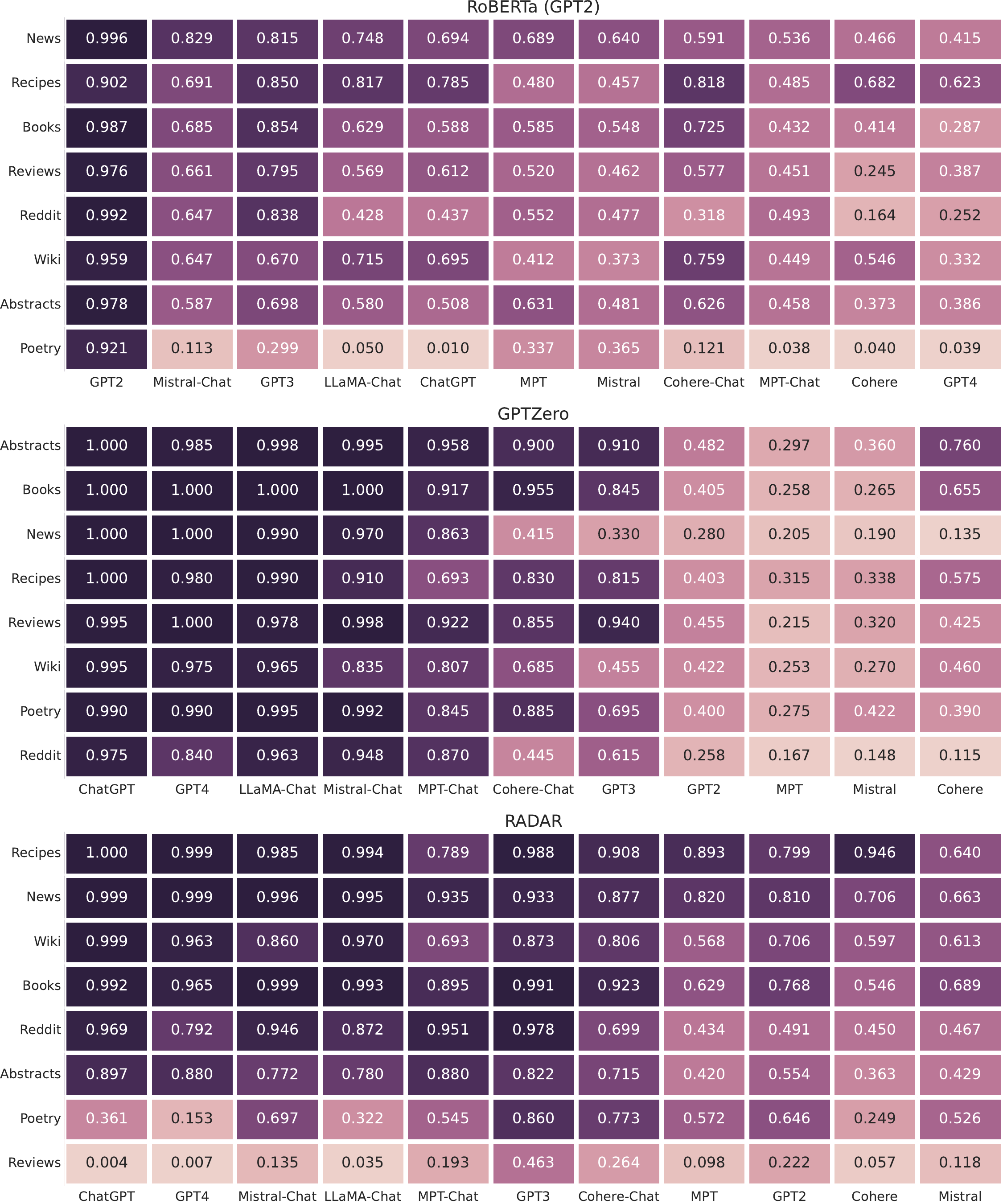}
    \caption{Extended heatmap of RoBERTa GPT2, GPTZero, and RADAR's performance across all models and domains in the RAID dataset. We see that the trends noted in Figure \ref{fig:heatmap} still hold.}
    \label{fig:app_binoculars_heatmap}
\end{figure*}

In Figure \ref{fig:app_binoculars_heatmap} we show the extended heatmaps from Figure \ref{fig:heatmap} in the main paper. We see that the trend holds that RoBERTa GPT2 is significantly better on GPT2 generations and that RADAR is uncharacteristically bad on IMDb Movie Reviews.

\begin{table*}[t]
    \small
    \centering
    \begin{tabular}{l|c|c|c|c|c|c|c|c}
    \toprule 
    &\textbf{News}&\textbf{Wiki}&\textbf{Reddit}&\textbf{Books}&\textbf{Abstracts}&\textbf{Reviews}&\textbf{Poetry}&\textbf{Recipes}\\
    \midrule
    RoBERTa-B GPT2&74.3&64.7&56.3&67.9&56.3&69.1&23.9&64.6\\
    RoBERTa-L GPT2&69.8&59.5&54.1&62.4&58.9&58.2&24.4&67.2\\
    RoBERTa-B CGPT&45.1&48.7&44.6&53.5&72.3&65.3&32.0&5.9\\
    RADAR&88.0&76.8&71.8&84.5&66.7&14.1&53.0&\textbf{88.5}\\
    \midrule
    GLTR&66.9&64.3&65.7&74.0&62.0&67.3&34.8&67.2\\
    FastDetectGPT&74.2&77.3&70.9&76.2&76.4&77.5&63.4&74.6\\
    LLMDet&39.8&32.6&39.6&37.1&18.0&33.1&30.7&48.1\\
    Binoculars&80.7&76.7&79.4&83.7&79.1&80.1&\textbf{81.0}&76.6\\
    \midrule
    GPTZero&58.1&62.8&57.0&71.4&74.9&70.5&69.5&67.6\\
    Originality&\textbf{88.4}&\textbf{83.2}&\textbf{85.0}&\textbf{90.4}&87.7&\textbf{87.3}&75.1&82.8\\
    Winston&72.4&54.9&68.9&70.7&\textbf{94.7}&72.9&64.3&68.9\\
    ZeroGPT(*)&72.2&70.6&65.1&73.3&60.3&68.6&50.0&63.7\\
    \bottomrule
    \end{tabular}
    \caption{Accuracy Score at FPR=5\% for detectors across different domains. We see that metric-based methods perform surprisingly well across domains and that detectors can perform surprisingly poorly on unseen domains.}
    \label{tab:per_domain_results}
\end{table*}

\subsection{Model Performance vs. Domain}
In Table \ref{tab:per_domain_results} we report the results of our 12 detectors across all different domains of generated text. We see that the metric-based methods such as Binoculars and FastDetectGPT generalize surprisingly well across domains. We also see that in general detectors perform well but occasionally perform surprisingly poorly.

\subsection{Detector Accuracy vs. Decoding Strategy}
In Table \ref{tab:decoding_full_results} we report the results of our evaluation broken up by category of model and by decoding strategy. Much like in Figure \ref{fig:decrep_barplot} from the main paper, we see that certain combinations of decoding strategies and models can cause detector accuracy to plummet unexpectedly. This raises serious concerns for the robust deployment of detectors.

\subsection{Detector Accuracy vs. Adversarial Attack}
In Table \ref{tab:adversarial_results} we report the full version of the results from Table \ref{tab:adversarial_comparison} in the main paper. We see similar trends, namely that certain adversarial attacks work better on certain detectors and that occasionally adversarial attacks actually improve detector performance rather than harm it. One notable inclusion here is the zero-width space attack, which seems to either cause detectors to assign all positive or all negative labels. Future work should investigate this phenomenon.

\begin{table*}[t]
    \small
    \centering
    \begin{tabular}{l|c|c|c|c|c|c|c|c|c|c|c|c}
    \toprule
    &\textbf{GPT2}&\textbf{GPT3}&\textbf{ChatGPT}&\textbf{GPT4}&\multicolumn{2}{c|}{\textbf{Cohere}}&\multicolumn{2}{c|}{\textbf{Mistral}}&\multicolumn{2}{c|}{\textbf{MPT}}&\textbf{Llama}&\textbf{Total}\\
    \midrule
    Chat? (Y/N)&\xmark&\xmark&\cmark&\cmark&\xmark&\cmark&\xmark&\cmark&\xmark&\cmark&\cmark&-\\
    \midrule
    R-B GPT2&84.0&74.7&65.4&42.4&42.9&61.1&45.7&65.9&49.2&45.9&68.7&59.1\\
    R-L GPT2&96.3&72.4&53.7&33.8&37.3&56.6&47.6&60.5&52.6&41.6&56.7&56.7\\
    R-B CGPT&36.7&54.2&66.1&30.0&31.3&49.6&17.7&69.9&17.8&53.3&70.1&44.8\\
    RADAR&64.7&88.6&82.1&76.0&51.4&77.2&54.1&83.6&58.0&76.5&78.5&70.9\\
    \midrule
    GLTR&66.6&85.1&81.4&53.7&54.2&67.4&49.1&75.4&35.4&52.5&81.6&62.6\\
    F-DetectGPT&72.1&95.4&96.1&73.9&84.7&85.1&58.2&81.3&45.3&57.1&94.0&73.6\\
    LLMDet&48.2&40.2&18.9&27.0&32.6&35.6&31.5&35.7&28.2&21.4&55.1&35.0\\
    Binoculars&68.9&\textbf{99.2}&\textbf{99.6}&91.9&\textbf{94.8}&\textbf{95.4}&62.3&91.7&45.2&70.8&97.6&79.6\\
    \midrule
    GPTZero&38.8&70.1&99.4&97.1&43.9&74.6&28.9&\textbf{95.6}&24.8&85.9&\textbf{98.5}&66.5\\
    Originality&\textbf{99.1}&98.2&98.2&89.9&78.9&90.6&\textbf{71.0}&95.5&\textbf{58.1}&76.2&94.7&\textbf{85.0}\\
    Winston&47.6&77.8&\textbf{99.6}&\textbf{98.8}&63.6&86.2&46.1&94.9&24.8&\textbf{79.2}&97.5&71.0\\
    ZeroGPT(*)&42.4&90.2&93.2&67.1&65.9&76.6&49.3&81.4&27.3&66.0&93.7&65.5\\
    \bottomrule
    \end{tabular}
    \caption{Accuracy at FPR=5\% for detectors on non-adversarial outputs of different models. We see that base models are more difficult to detect than their chat fine-tuned counterparts and that metric-based methods show impressive cross-model generalization. Asterisks (*) indicate that the detector was unable to achieve the target FPR.}
    \label{tab:per_model_results}
\end{table*}

\begin{table*}[t]
    \small
    \centering
    \begin{tabular}{l|c|c|c|c|c|c|c|c|c|c|c|c}
    \toprule
    &\textbf{None}&\textbf{AS}&\textbf{AD}&\textbf{HG}&\textbf{IP}&\textbf{NS}&\textbf{PP}&\textbf{MS}&\textbf{SYN}&\textbf{ULS}&\textbf{WSA}&\textbf{ZWS}\\
    \midrule
    RoB-B GPT2&59.1&55.6&37.1&7.6&56.9&55.9&68.9&43.8&71.5&18.8&45.2&99.9\\
    RoB-L GPT2&56.7&52.4&33.2&21.3&55.1&51.7&72.9&39.5&79.4&19.3&40.1&99.9\\
    RoB-B CGPT&44.8&43.3&38.0&0.0&5.2&44.3&49.2&42.1&39.6&31.7&0.1&0.0\\
    RADAR&70.9&70.8&67.9&59.3&73.7&71.0&67.3&69.5&67.5&70.4&66.1&82.2\\
    \midrule
    GLTR&62.6&61.2&52.1&24.3&61.4&59.9&47.2&59.8&31.2&48.1&45.8&97.2\\
    F-DGPT&73.6&71.6&64.7&51.4&72.0&68.2&71.8&70.7&34.0&60.4&64.4&98.9\\
    LLMDet&35.0&33.9&27.4&40.6&27.2&33.8&28.5&32.7&27.3&23.4&4.4&27.1\\
    Binoculars&79.6&78.2&74.3&37.7&71.7&77.1&80.3&78.0&43.5&73.8&70.1&99.1\\
    \midrule
    GPTZero&66.5&64.9&61.0&66.2&66.2&65.8&64.0&65.1&61.0&56.5&66.2&66.2\\
    ZeroGPT&65.5&65.4&59.7&82.4&64.9&64.7&46.7&64.7&18.8&54.5&64.2&48.0\\
    Originality&85.0&83.6&71.4&9.3&85.1&86.0&96.7&78.6&96.5&75.8&84.9&4.9\\
    Winston&71.0&68.9&66.9&26.3&69.8&69.0&52.6&67.5&63.6&56.8&46.8&25.0\\
    \bottomrule
    \end{tabular}
    \caption{Accuracy Score at FPR=5\% for all detectors across different adversarial attacks. Abbrevations are: AS: Alternative Spelling, AD: Article Deletion, HG: Homoglyph, IP: Insert Paragraphs, NS: Number Swap, PP: Paraphrase, MS: Misspelling, SYN: Synonym Swap, ULS: Upper Lower Swap, WSA: Whitespace Addition, ZWS: Zero-Width Space Addition}
    \label{tab:adversarial_results}
\end{table*}

\section{Example Generations}
\label{app:examples}
In Table \ref{tab:example_generations} and \ref{tab:adversarial_attack_examples} we provide example outputs for each generative model and adversarial attack. We see that different generative models have significantly differing styles, underscoring the difficulty of the detection problem.

\begin{table*}
\centering 
\small
\begin{tabular}{p{0.11\linewidth}|p{0.83\linewidth}} 
\toprule
\multicolumn{2}{p{0.94\linewidth}}{\textbf{Domain}: Reddit, \textbf{Title}: VW album track list/lyric sheet inquiry? (\textbf{ID}: 25fb9f14-c7c5-4e76-95ac-bf149780f30d)}\\
\midrule
\multicolumn{2}{p{0.94\linewidth}}{\textbf{Prompts}: (\textbf{Chat}) Write just the body of a Reddit post titled "VW album track list/lyric sheet inquiry?".\newline\quad(\textbf{Non-Chat}): The following is the full text of a post titled "VW album track list/lyric sheet inquiry?" from reddit.com}\\
\midrule
\textbf{Original Reddit Post}\newline (Human)&I bought a "Modern vampires of the city" on vinyl recently, and I was wondering if anyone had information on the nature of the track list? Its on the backside of the poster, and its a normal tracklist and lyric sheet except there's some things circled and underlined and stuff. I looked it up, and I thought origionally that it was supposed to look like a playbook, like football, but it looks like the marks are intentional to the numbers and stuff so then I thought, maybe it was them planning their album? And they just decided to add some flair to the track list? Either way, im curious, and a quick Google search didn't work, so I figured you guys would know better than me. I would love any input I could get\\
\midrule
\textbf{GPT-2}&"Hi, this is a thread for people of any ability to help me with a query related to the following track: "Cherries On The Wind" from the album 'Volkswagen', which was released July 2008 to August 2009. I have a few questions about all of them (my two biggest questions are 'Who sings it?' and 'Who wrote it and why?'):

(1) Is the singer named "Cherries" a female? If not, is it a boy, female or something else?

(2) Is it a female lead or male supporting singer in the band?

(3) Who writes the lyrics which are written in both German and English, in the style of a pop song? Is it a German name or is it the name of a song artist, musician or songwriter?

(4) Is there a female vocalist in the band? If so, can you shed some light on the singer? Are there German solo projects by her? If so, can you shed some light on the singer? Are there German solo projects by her? http://www.youtube.com/watch?v=jY9FyMwKvCk\&feature=fvrs

If you need to ask more question, just add them to the top of the post and I'll answer. Thanks much!

2 people have voted this message helpful

Lyrics: https://www.youtube.com/watch?v=fSdWJ4Wy7k0\\
\midrule
\textbf{GPT-3}&Hey all, I'm wondering if anyone might be able to help me out with something. I'm trying to track down a copy of the lyric sheet/track list for the album "Volkswagen." I've been able to find a few images online, but they're all partial and/or low quality. If anyone has a high quality scan or knows where I can find one, I would really appreciate it! Thanks in advance!\\
\midrule
\textbf{ChatGPT}&Hello everyone, I recently got my hands on a vinyl copy of VW's latest album and I am absolutely loving it! The production, the melodies, and the lyrics are just top-notch. I've been obsessively listening to it and can't get enough. However, I was wondering if anyone has access to or knows where I can find the track list and lyric sheet for the album? I've been searching online but haven't had much luck so far. I would really love to dive deeper into the lyrics and understand the meaning behind some of the songs. 
If anyone can point me in the right direction or has any information, I would greatly appreciate it. Thanks in advance!\\
\midrule
\textbf{GPT-4}&Hello fellow VW fans! I've recently got my hands on a vintage vinyl copy of one of the older Vampire Weekend albums. However, it didn't come with a track list or lyric sheet. It's proving to be a bit of a task to identify some of the tracks and I'd love to sing along to my favourites. If anyone could point me to where I might find a track list and/or a place where I can find the lyrics to the songs, I would greatly appreciate it. Let me know or feel free to message me directly. Thanks for your help, guys!\\
\midrule
\textbf{Mistral}&> This has been bugging me for a bit. I never got my copy of "VW" (or "VW EP" - whichever you want to call it) so I never had a chance to look at the lyric sheet. I know there's an extended version on iTunes, but I haven't the funds to purchase it since the EP and "TNWT" came out at the same time. But I would like to know the lyrics to at least "Wonderful Life" (for which I know there's no lyrics of), "The New Wave", "Song of Our Hits", "What If God Was One Of Us", and "The Hits Just Keep Coming". I'd like to know how the song "What If God Was One Of Us" fits with the storyline (or is it related at all?), as well as the whole thing behind "What Would She Say?", and the ending song that has to be either a spoiler for "VW" ("I Am the Wind"), or something relating to "TNWT" (I think that line from the "TNWT" lyric sheet about a "new power" in the verse between "The Hits Just Keep Coming" and "Song For Our Hits" is what I'm thinking of)\\
\midrule
\textbf{Mistral Chat}&Hey everyone, I just got into The Velvet Underground and I've been a huge fan of their work ever since. I've been doing a deep dive into their discography and I was wondering if anyone had a track list or lyrics sheet for the album "VW"? I can't find it anywhere online and it's driving me crazy. Any help would be greatly appreciated! Thanks in advance!\\
\midrule
\textbf{Cohere}&VW album track list/lyric sheet inquiry? Hello! I was wondering if anyone had scanned/digitized the track list or lyric sheet from the VW album (oil diffuser). I know it's floating around here somewhere, but damned if I can find it. Thanks!\\
\midrule
\textbf{Cohere Chat}&Check this out! Been waiting for this for ages!! Hey all – I've recently fallen in love with VW's latest album (thank you Reddit for the recommendation!) but I cannot for the life of me find the track list or lyrics anywhere. The songs are amazing but still quite new for me, so I have no clue what the lyrics are and wanna know!!! Does anyone have a track list/ lyric sheet that they could share? Any help is appreciated!\\
\midrule
\end{tabular}
\end{table*}

\begin{table*}
\centering 
\small
\begin{tabular}{p{0.11\linewidth}|p{0.83\linewidth}} 
\midrule
\textbf{MPT}&I read this on Reddit and think it bears repeating and expanding upon on a site that VW followers frequent, and I figured it behooved me to get it out to you guys ASAP
The first batch of test pressings for the new Volcano Witch album, which was recorded during a recent trip to Chicago by VW member John McEntire, will be arriving very soon. Before this happens, the band wanted to know whether or not our fans will be interested in a test pressing lyric sheet. This is a special edition (limited number) item that will only be made available to those supporters that were very instrumental to the creation \& completion of the new VW album. The test pressing lyric sheet will feature all 10 tracks with full handwritten lyrics by John, who also took the time to hand write lyrics to two new never heard before VW songs, which will only be available on the lyric sheet \& test pressings of the album. If our followers are interested in this special edition item please submit a screenshot of the below image with the word “YES” placed in the comment section of this post. If you’re unable to submit a screenshot simply place the word “YES” in the comment section of this post \& I’ll make sure you’re added to list for the special test pressing lyric sheet. All proceeds from the test pressing will be donated to The Center For Discovery, the school my daughter attends, in her honor…
In the meantime, while the band waits to hear from their fans, we’re giving you the chance to be one of 100 lucky winners to win a signed “Storm Lord” poster that was created by the band’s longtime collaborator Eric Adrian Lee, who also designed \& made all the artwork for the new Volcano Witch album. To enter yourself into this giveaway all you have to do is share this post. That’s it, simple as that. We’ll pick the lucky winners on October 2nd \& then post the giveaway winners on our social media. So…are you interested in the special test pressing lyric sheet?
Source: http://redd.it/30fsl1
Posted at 10:47 PM in Volcano Witch | Permalink | Comments (0)
Volcano Witch - "A.M. (Live at the Empty Bottle)"\\
\midrule
\textbf{MPT Chat}&Hey everyone! I'm a big fan of VW and I was wondering if anyone knows where I can find the album track list and lyric sheet for their latest album? I've been searching online but can't seem to find it anywhere. Any help would be much appreciated! Thanks in advance!\\
\midrule
\textbf{Llama Chat}&Hey fellow Redditors, I'm hoping someone can help me out with a little inquiry I have about the Velvet Underground's album "The Velvet Underground \& Nico". I've been trying to find a track list/lyric sheet for the album that includes the bonus tracks from the CD reissue, but I can't seem to find one anywhere. I've searched high and low on Google, but all the track lists I come across only include the original track list for the album, without the bonus tracks. I've also checked the album's Wikipedia page and various other websites, but no luck. Does anyone know of a resource or website that has the complete track list/lyric sheet for the album, including the bonus tracks? I'd really appreciate it if someone could point me in the right direction. Thanks in advance for the help, and sorry if this has been asked before (I searched the subreddit but didn't find anything).\\
\bottomrule
\end{tabular}
\caption{Output generations from each model in RAID for the Reddit post titled ``VW album track list/lyric sheet inquiry?'' using random sampling (temp=1, p=1) and no repetition penalty. Prompts used are listed in Table \ref{tab:generation_prompts}}
\label{tab:example_generations}
\end{table*}

\begin{table*}
\centering 
\small
\begin{tabular}{p{0.11\linewidth}|p{0.83\linewidth}}
\toprule
\multicolumn{2}{p{0.94\linewidth}}{\textbf{Domain}: Reviews, \textbf{Model}: Cohere, \textbf{Decoding}: Sampling (No Penalty), (\textbf{ID}: b0aa73c4-ff31-4a43-9472-8fe7a85e2754)\newline
\textbf{Prompt}: The following is the full text of a review for the movie "Fast Five" from IMDb.com:}\\
\midrule
\textbf{Original}\newline (No Attack)&This is probably my favorite action movie of 2011, can't wait to see Fast and Furious 6. Bringing back Vin Diesel and the late Paul Walker was a great idea, making this franchise better than ever. The action was great and suspenseful, and the new cast members were amazing. This is a must-see.\\
\midrule
\textbf{Alternative Spelling}&This is probably my {\color{blue}favourite} action movie of 2011, can't wait to see Fast and Furious 6. Bringing back Vin Diesel and the late Paul Walker was a great idea, making this franchise better than ever. The action was great and suspenseful, and the new cast members were amazing. This is a must-see.\\
\midrule
\textbf{Article\newline Deletion}&This is probably my favorite action movie of 2011, can't wait to see Fast and Furious 6. Bringing back Vin Diesel and the late Paul Walker was {\color{blue}\sout{a}} great idea, making this franchise better than ever. The action was great and suspenseful, and {\color{blue}\sout{the}} new cast members were amazing. This is a must-see.\\
\midrule
\textbf{Homoglyph}&Th{\color{blue}i}s {\color{blue}i}s {\color{blue}p}r{\color{blue}o}b{\color{blue}a}bl{\color{blue}y} m{\color{blue}y} f{\color{blue}a}v{\color{blue}o}r{\color{blue}i}te {\color{blue}a}{\color{blue}c}t{\color{blue}i}{\color{blue}o}n m{\color{blue}o}v{\color{blue}i}{\color{blue}e} {\color{blue}o}f 2011, {\color{blue}c}{\color{blue}a}n't w{\color{blue}a}{\color{blue}i}t t{\color{blue}o} s{\color{blue}e}{\color{blue}e} F{\color{blue}a}st {\color{blue}a}nd Fur{\color{blue}i}{\color{blue}o}us 6. {\color{blue}B}r{\color{blue}i}ng{\color{blue}i}ng b{\color{blue}a}{\color{blue}c}k V{\color{blue}i}n D{\color{blue}i}{\color{blue}e}s{\color{blue}e}l {\color{blue}a}nd th{\color{blue}e} l{\color{blue}a}t{\color{blue}e} P{\color{blue}a}ul W{\color{blue}a}lk{\color{blue}e}r w{\color{blue}a}s {\color{blue}a} gr{\color{blue}e}{\color{blue}a}t {\color{blue}i}d{\color{blue}e}{\color{blue}a}, m{\color{blue}a}k{\color{blue}i}ng th{\color{blue}i}s fr{\color{blue}a}n{\color{blue}c}h{\color{blue}i}s{\color{blue}e} b{\color{blue}e}tt{\color{blue}e}r th{\color{blue}a}n {\color{blue}e}v{\color{blue}e}r. Th{\color{blue}e} {\color{blue}a}{\color{blue}c}t{\color{blue}i}{\color{blue}o}n w{\color{blue}a}s gr{\color{blue}e}{\color{blue}a}t {\color{blue}a}nd sus{\color{blue}p}{\color{blue}e}ns{\color{blue}e}ful, {\color{blue}a}nd th{\color{blue}e} n{\color{blue}e}w {\color{blue}c}{\color{blue}a}st m{\color{blue}e}mb{\color{blue}e}rs w{\color{blue}e}r{\color{blue}e} {\color{blue}a}m{\color{blue}a}z{\color{blue}i}ng. Th{\color{blue}i}s {\color{blue}i}s {\color{blue}a} must-s{\color{blue}e}{\color{blue}e}.\\
\midrule
\textbf{Insert\newline Paragraphs}&This is probably my favorite action movie of 2011, can't wait to see Fast and Furious 6. Bringing back Vin Diesel and the late Paul Walker was a great idea, making this franchise better than ever.\newline
{\color{blue}[\P]}

The action was great and suspenseful, and the new cast members were amazing. This is a must-see.\\
\midrule
\textbf{Number Swap}&This is probably my favorite action movie of {\color{blue}0346}, can't wait to see Fast and Furious 6. Bringing back Vin Diesel and the late Paul Walker was a great idea, making this franchise better than ever. The action was great and suspenseful, and the new cast members were amazing. This is a must-see.\\
\midrule
\textbf{Paraphrase}&This is {\color{blue}surely the best of all the} action {\color{blue}films in} 2011. {\color{blue}I'm really looking forward to} Fast \& Furious 6. {\color{blue}I loved that they brought back} Vin-Diesel and the late Paul-Walker, {\color{blue}it made the series} better than ever. The action was suspenseful {\color{blue}and exciting} and the new {\color{blue}actors} were all {\color{blue}great}. This is a must-see.\\
\midrule
\textbf{Misspelling}&This is probably my favorite action movie of 2011, can't wait to see Fast and Furious 6. Bringing back Vin Diesel and the late Paul Walker was a {\color{blue}grat} idea, making this franchise better than ever. The action was {\color{blue}grate} and suspenseful, and the new cast members were amazing. This is a must-see.\\
\midrule
\textbf{Synonym Swap}&This is {\color{blue}also} my favourite action {\color{blue}film} of 2011, can't wait to see Fast and Furious 6. {\color{blue}Taking down} Vin Diesel and the late {\color{blue}Scott Davis} was a {\color{blue}good thing}, {\color{blue}doing} this franchise better than ever. The action was {\color{blue}good} and suspenseful, and the new cast members were amazing. This is a must-see.\\
\midrule
\textbf{Upper Lower}&This is probably my favorite action movie of 2011, can't wait to see Fast and {\color{blue}f}urious 6. Bringing back Vin Diesel and the late Paul {\color{blue}w}alker was a great idea, making this franchise better than ever. The action was great and suspenseful, and the new cast members were amazing. This is a must-see.\\
\midrule
\textbf{Whitespace}&This is probably my favorite{\color{blue}[ ]} action movie of 2011, can't wait to see Fast and{\color{blue}[ ]} Furious 6. Bringing{\color{blue}[ ]} back Vin Diesel{\color{blue}[ ]} and the late Paul Walker was a great idea,{\color{blue}[ ]} making this{\color{blue}[ ]} franchise better{\color{blue}[ ]} than ever. The action was great and suspenseful, and the new cast members{\color{blue}[ ]} were amazing. This is a must-see.\\
\midrule
\textbf{Zero-Width Space}&T{\color{blue}[U+200B]}h{\color{blue}[U+200B]}i{\color{blue}[U+200B]}s{\color{blue}[U+200B]} i{\color{blue}[U+200B]}s{\color{blue}[U+200B]} {\color{blue}[U+200B]}p{\color{blue}[U+200B]}r{\color{blue}[U+200B]}o {\color{blue}[U+200B]}b{\color{blue}[U+200B]}a{\color{blue}[U+200B]}b{\color{blue}[U+200B]}l{\color{blue}[U+200B]}y{\color{blue}[U+200B]} {\color{blue}[U+200B]}m{\color{blue}[U+200B]}y{\color{blue}[U+200B]} {\color{blue}[U+200B]}f{\color{blue}[U+200B]}a{\color{blue}[U+200B]}v{\color{blue}[U+200B]}o{\color{blue}[U+200B]}r{\color{blue}[U+200B]}i{\color{blue}[U+200B]}t{\color{blue}[U+200B]}e{\color{blue}[U+200B]} {\color{blue}[U+200B]}a{\color{blue}[U+200B]}c{\color{blue}[U+200B]}t{\color{blue}[U+200B]}i{\color{blue}[U+200B]}o{\color{blue}[U+200B]}n{\color{blue}[U+200B]} {\color{blue}[U+200B]}<...>\\
\bottomrule
\end{tabular}
\caption{Example outputs for each adversarial attack in RAID when applied to the IMDb movie review for ``Fast Five'' generated by Cohere using random sampling (temp=1, p=1) and no repetition penalty. Blue color indicates the portion of the text that was changed by the attack. Detailed descriptions of each attack along with their effective attack surfaces are listed in Appendix \ref{app:adversarial}.}
\label{tab:adversarial_attack_examples}
\end{table*}

\end{document}